\documentclass[10pt,letterpaper]{article}
\usepackage[top=0.85in,left=2.75in,footskip=0.75in,marginparwidth=2in]{geometry}

\usepackage{nameref} 

\usepackage[utf8]{inputenc}
\usepackage[T1]{fontenc}
\usepackage{enumitem}
\usepackage{lmodern}

\usepackage{hyperref}

\usepackage[normalem]{ulem}

\usepackage{amsmath, amssymb, amsfonts}

\usepackage[right, mathlines]{lineno}

\usepackage{microtype}
\DisableLigatures[f]{encoding = *, family = * }

\raggedright
\usepackage{changepage}

\usepackage[aboveskip=0.8pt,labelfont=bf,labelsep=period,singlelinecheck=off]{caption}

\makeatletter
\renewcommand{\@biblabel}[1]{\quad#1.}
\makeatother

\usepackage{lastpage,fancyhdr,graphicx}
\usepackage{epstopdf}
\fancyheadoffset[L]{2.25in}
\fancyfootoffset[L]{2.25in}

\usepackage{xcolor}
\definecolor{captionColour}{rgb}{0.21, 0.27, 0.31}

\usepackage{graphicx}

\usepackage{sidecap}
\renewcommand{\figurename}{Fig.  } 
\newcommand{\eqname}{} 

\usepackage{wrapfig}
\usepackage[pscoord]{eso-pic}
\usepackage[fulladjust]{marginnote}
\reversemarginpar

\usepackage{amsmath, amssymb, amsfonts}
\usepackage{mathtools}

\newcommand{\defeq}{\vcentcolon=}

\usepackage{amsthm}
\theoremstyle{plain}
\newtheorem{theorem}{Theorem}

\theoremstyle{definition}
\newtheorem{definition}{Definition}

\theoremstyle{remark}

\DeclareMathOperator*{\E}{\mathbb{E}}

\renewcommand{\S}{\mathcal{S}}
\newcommand{\A}{\mathcal{A}}
\newcommand{\G}{\mathcal{G}}

\renewcommand{\L}{\mathcal{L}}

\usepackage{scalerel}

\newcommand{\Fen}{\mathcal{F}}
\newcommand{\pif}{{\scaleto{\pi}{4pt}_{\scaleto{\Fen}{3pt}}}}
\newcommand{\pifi}[1]{{\scaleto{\pi}{4pt}_{\scaleto{\Fen}{3pt}}^{\scaleto{(#1)}{5pt}}}}

\newcommand{\piv}{{\pi_{\scaleto{V}{3pt}}}}

\newcommand{\Itg}{\scaleto{\Im}{8pt}}
\newcommand{\ItgD}{\Itg_{\scaleto{D}{3pt}}}
\newcommand{\ItgDpif}{{\Itg^{\pif}_{\scaleto{D}{3pt}}}}
\newcommand{\ItgDpifi}[1]{{\Itg^{\pifi{#1}}_{\scaleto{D}{4pt}}}}

\DeclareMathOperator*{\argmax}{arg\,max\,}
\DeclareMathOperator*{\argmin}{arg\,min\,}

\usepackage{stmaryrd,graphicx}
\makeatletter
\newcommand{\fixed@sra}{$\vrule height 2\fontdimen22\textfont2 width 0pt\shortuparrow$}
\newcommand{\shortarrow}[1]{%
  \mathrel{\text{\rotatebox[origin=c]{\numexpr#1*45}{\fixed@sra}}}
}
\makeatother

\usepackage{silence}
\usepackage{placeins}

\usepackage{algorithm}
\usepackage{algorithmicx}
\usepackage[noend]{algpseudocode}

\usepackage{xargs}

\usepackage[numbers, sort&compress, square]{natbib}

\begin{document}

\vspace*{0.35in}

\begin{flushleft}
{\Large
\textbf\newline{A space of goals: the cognitive geometry of informationally bounded agents}
}
\newline
\\
Karen Archer\textsuperscript{1,*},
Nicola Catenacci Volpi\textsuperscript{1},
Franziska Bröker\textsuperscript{2,3},
Daniel Polani\textsuperscript{1}
\\
\bigskip
{1} Adaptive Systems Group, Department of Computer Science, University of Hertfordshire, Hatfield, United Kingdom
\\
{2} Gatsby Computational Neuroscience Unit, University College London, London, United Kingdom
\\
{3} Max Planck Institute for Biological Cybernetics, Tübingen, Germany
\\
\bigskip
* karen.archer@gmail.com

\end{flushleft}

\section*{Abstract}

Traditionally, Euclidean geometry is treated by scientists as a priori and objective. However, when we take the position of an agent, the problem of selecting a best route should also factor in the abilities of the agent, its embodiment and particularly its cognitive effort. In this paper we consider geometry in terms of travel between states within a world by incorporating information processing costs with the appropriate spatial distances. This induces a geometry that increasingly differs from the original geometry of the given world as information costs become increasingly important. We visualise this \textit{``cognitive geometry''} by projecting it onto 2- and 3-dimensional spaces showing distinct distortions reflecting the emergence of epistemic and information-saving strategies as well as pivot states. The analogies between traditional cost-based geometries and those induced by additional informational costs invite a generalisation of the notion of geodesics as cheapest routes towards the notion of \textit{infodesics}. In this perspective, the concept of infodesics is inspired by the property of geodesics that, travelling from a given start location to a given goal location along a geodesic, not only the goal, but all points along the way are visited at optimal cost from the start.

\textbf{Keywords}: information-regularised MDPs, decision sequences, constrained information processing, geometry, cognitive load


\section{Introduction}

Traditional Reinforcement Learning (RL) \citep{Sutton2018} is often centred around agents moving towards a specific goal.  The problem of designing agents that can address multiple concurrent objectives at the same time has recently become of pivotal importance in RL   \citep{liu2014multiobjective, yang2019generalized}.

While start and goal states are part of the same world, the resulting
policies are often merely collected in the form of a ``catalogue'' for
the different start/goal combinations. Here we wish to start from a
fundamentally geometric position: we posit that in a world where the
agent incurs movement costs for each action (negative rewards only),
and trajectories implement directed shortest routes between a start
and a goal point, these become a part of a geometry. Such a structure puts these different trajectories and the policies used to achieve them in relation with each other and allows us to systematically draw conclusions about their commonalities, what the space between these tasks looks like and how to switch between tasks or trajectories, and, finally, how to generalise solutions. The best-known form of such a structure is Euclidean geometry.

Classical Euclidean geometry is often considered as a ``static''
concept. However, one can alternatively interpret it as the
entirety of locations of a set under consideration, together with the
straight lines defining the shortest routes of travel between these
points, and the associated directions determined by such straight
lines.  In the field of cognitive science, this Euclidean notion of lines connecting bounding points has been used in the theory of Conceptual Spaces \citep{Gardenfors:2000-conceptual-spaces-geometry-of-thought} to geometrically represent knowledge. It introduces a conceptual space which uses geometric and topological ideas to represent quality dimensions which enable one to compare concepts such as weight, colour, taste, etc. Natural categories are convex regions in conceptual spaces, in that if $x$ and $y$ are elements of a category, and if $z$ is between $x$ and $y$, then z is also likely to belong to the category. This convexity allows the interpretation of the focal points of regions as category prototypes.  Analogous to straight lines in Euclidean space, the distance in between concepts is defined in terms of these prototypes through conceptual similarity \citep{Tversky1977:features-of-similarity} which expresses their similarity as a linear combination of their common and distinctive features. 

In contrast to such a concept of ``in-betweenness'', in traditional
geometry, one has furthermore the alternative option of navigating by
choosing an initial direction for a route which then is subsequently
maintained in a consistent fashion. Using a compass an agent can
maintain a bearing, say, a strict northern course, by selecting the
desired direction and then strictly keeping its current orientation.
This can be used to reach a goal, or even multiple goals, optimally
whenever this fixed direction happens to coincide with one or more
desired shortest routes. This concept generalises into Riemannian
geometry by replacing the straight lines with geodesics which
represent a kind of consistent directionality on a manifold. This
assumes that direction-constant and shortest routes coincide,
otherwise they need to be treated separately.

While Riemannian geometry still respects the symmetry of distance
between locations, one can generalise this to direction-dependent
routes in terms of a \emph{Finsler geometry} \citep{Wilkens1995}, where such
distances are no longer symmetric between two given points; this
models, for instance, non-reversible energy expenditure, altitude
traversal, or accounting for the ease of terrain. This can be
elegantly expressed as RL problems which carry only positive cost (or
negative reward, our convention throughout the paper). In this context
the RL policy is a decision or behaviour strategy employed by an agent
to solve a task. In this paper, we will consider, amongst other, the
limit case of an open-loop policy with optimal costs; we will argue
that this can be interpreted as a generalisation of selecting and
following a cardinal direction  by a compass while keeping
the orientation fixed, which is a state-independent behaviour. More precisely, once an open-loop policy, i.e.\
a bearing has been determined, using the same action or action
distribution selected in every state, the agent will proceed without
further deliberation or intake of information about the current state,
i.e.\ in an open-loop manner. We interpret this as corresponding to
strictly following the set direction, without adjustment.

When considering behaviour, it is traditionally hypothesised that
behavioural preferences of rational agents can be characterised by a
suitably chosen utility function $U$ that assigns a numerical value to
the possible behaviour outcomes, and the agents act as to maximise the
expected value of this utility
\citep{Von_Neumann1944:expected-utility}. System deviations from
rational behaviour are known as cognitive biases which result from the
use of fast but fallible cognitive strategies known as heuristics
\citep{Lieder2020:resource-rational-cognition}. Agents have finite
cognitive resources. Thus, it is unrealistic to assume that agents would have a unique heuristic for every particular situation. Resource-rationality aims to provide a framework to account for heuristics and biases, such that expected utility maximisation is subject to both computational costs and cognitive limitations. In this context, acting optimally is defined as the decision strategy, or policy $\pi$, which maximises the difference between the expected utility and the cost of the policy \citep{Ho2022:cognitive-decisions-robots} according to $\pi^* = {\arg \max}_\pi \left(\E_\pi[U] - \operatorname{Cost}(\pi)\right)$. These costs include the cost of computation \citep{Griffiths2015:cost-of-computation} and the cost of control \citep{Shenhav2013:expected-utility-theory}. The cost for the cognitive processing required to operate a given policy has also been measured in terms of information-theoretic functionals \citep{Ortega2013, Rubin2012, Larsson2017, Zenon2019-information-theoretic-cognitive-costs}. In the present paper, we are interested in the geometries emerging
from the incorporation of such informational cognitive costs
\citep{catenacci2020space}. 

In two-dimensional continuous Euclidean spaces, the
optimal routes consist of straight lines which also form shortest
paths between two points. On a sphere, optimal routes lie on grand
circles connecting two points; furthermore, souch routes can be
naturally extended beyond their boundary points, on the sphere, to
full grand circles. The latter form \emph{geodesics} which no longer
necessarily constitute overall shortest routes, however, they are still \textit{direction-preserving curves}, the generalisation of the Euclidean straight line \citep{Pressley2010}. 
We will generally consider geometry in
terms of locations (points) and optimal routes from a start location
to another.  Such routes are characterised by a \emph{distance} (in
our RL-based convention, negative cumulated rewards) and a
\emph{direction}.

Rethinking RL in terms of geodesics means,  in some cases,
splitting the resulting trajectory at a subgoal $s'$, such that the agent can
visit $s'$ en route to the final goal $g$ without any extra
informational costs. Furthermore, instead of a
cost-optimal route between two states $s$ and $g$, one can instead consider
the starting state $s$ and a generalised ``direction'' towards $g$,
which in our case is reinterpreted as an open-loop policy.

Incorporating cognitive cost we can additionally ask how such a
geometry would ideally be organised and represented in an
informationally limited agent. We note that this perspective, while
carrying some similarity to the questions of transfer learning
\citep{taylor2009transfer}, is still markedly distinct. No symmetry or
equivariance is implied, only operational and cognitive distances
(costs) and their associated policies --- in particular, there is no
reason to assume that one type of strategy carries over to another
region. Of course, if there is some spatial coherence across regions,
one can expect some relevant overlap of trajectory structures between them.

We can consider our agent starting at a given point and
proceeding unchanged to ``move in one direction'', with a fixed given
behaviour or else, with a strategy optimal for a particular problem of
traveling from a state $s$ to $g$. Similarly to the pure cost problem,
we ask which goals in addition to the original one will be optimally
reached following this fixed behaviour or a least-cost route. Such
generalised geometries with their directional or minimal cost routes
lead us to envisage connections in the space that enable agents to
adapt policies to goals beyond our current goal. Whenever the
infodesic contains more than just the original start and end state,
this means that multiple goals can be reached without redirection or
extra costs (or with costs that do not exceed a given relaxation
threshold). In the case where we consider minimal cost routes and
cognitive costs incorporating the cost of processing information, we will
also speak of \emph{infodesics}. In these, constraining information
processing can thus trade in nominally shorter but more complicated
routes against longer, but simpler and safer ones
\citep{Rubin2012,Larsson2017}, which effectively imposes a distortion
on the geometry of a task space.

To investigate this further, we define \textit{Decision Information} to quantify the amount of information processed by an agent to execute a policy to navigate from a starting location to the goal state.    Here, we use an information-theoretic free energy principle \citep{Mitter2003, Tishby2011:information-to-go, Rubin2012, Ortega2013, Larsson2017
}, which induces a notion of informational distance and endows the decision-making problem with a qualitative geometrical interpretation.

We now introduce our formalism. After describing the simulations implemented, we present and discuss the results with suggestions for future avenues of research.

\section{Methods}
We seek to model an agent's behaviour and  the trade-off between information processing and performance.   Concretely, here we model discrete gridworlds as a discrete Markov Decision Process (MDP). Critically, we interpret action choice in a state as the discrete analogue of geometrical ``direction'' selection in that state.

\subsection{Markov Decision Process (MDP) Framework}

We model the system interactions between the environment and an agent ensuing from a sequence of decisions at discrete time steps for $t \in 0, \dots, T$ using an undiscounted MDP \citep{Puterman1994} defined by the tuple $\left <\S, \A, r, P\right >$.  We assume throughout that the trajectory is episodic, in particular, that the probability for trajectories generated by optimal policies to terminate at a goal state is one.   At time $t$, the state of the environment is $s_t \in \S$ ($\S$ denoting the set of states with random variable $S \in \mathcal{S}$, and similarly for the set of actions $\A$, with random variable $A \in \A$).  The policy $\pi(a_t|s_t)$ denotes the conditional probability distribution $\Pr\{A=a_t| S=s_t\}$ which defines a stochastic choice of actions in each state $s_t$ at time $t$.  Note that, as appropriate, we will drop the temporal index $t$ of $s$ and $a$ to emphasise that the policy, transition probability and reward function do not depend on time, but only on the states involved.

The dynamics of the MDP is modelled using state-action probability matrices $P^{a}_{ss'}$ which define the distribution of the successor state ($s_{t+1}$ is denoted as $s'$ where unambiguous, with random variable $S'\in \S$) given a current state and action, according to $P^{a}_{ss'} \equiv p(s' |s, a) \defeq \Pr\{S' = s'|S=s, A = a \} $.  As a consequence of each action $a_t \in \A$, the agent receives a reward $r(s', s, a)\leq 0$ in the subsequent time step (all the rewards from transient states are here chosen to be $-1$) and we refer to such an MDP as ``cost-only''.   Goal states, denoted by $g \in \S$, are absorbing, i.e.\ $p(s'=g \vert s=g, \cdot) = 1$, and $r(s'=g,s=g, \cdot)=0$, therefore once the agent reaches a goal, no further (negative) rewards are incurred.  The value function $V^{\pi}(s)$ represents the expected return of an agent when starting in a given state $s$ and following policy $\pi$ \citep{Sutton2018}.  Equation \eqref{eq:value-bellman} shows the value function in the form of a recursive Bellman equation which expresses the relationship between the value of a state and the values of its successor states, see \citep[section 3.5]{Sutton2018} and \ref{S-value-function} in the supplementary material.
\begin{equation} V^{\pi}(s) =
    \E_{\pi(a|s)p(s'|s, a)}\left[ r(s', s, a) +
      V^{\pi}(s')\right] \label{eq:value-bellman}
\end{equation}

\begin{figure}[!ht]
\includegraphics[width=\textwidth]{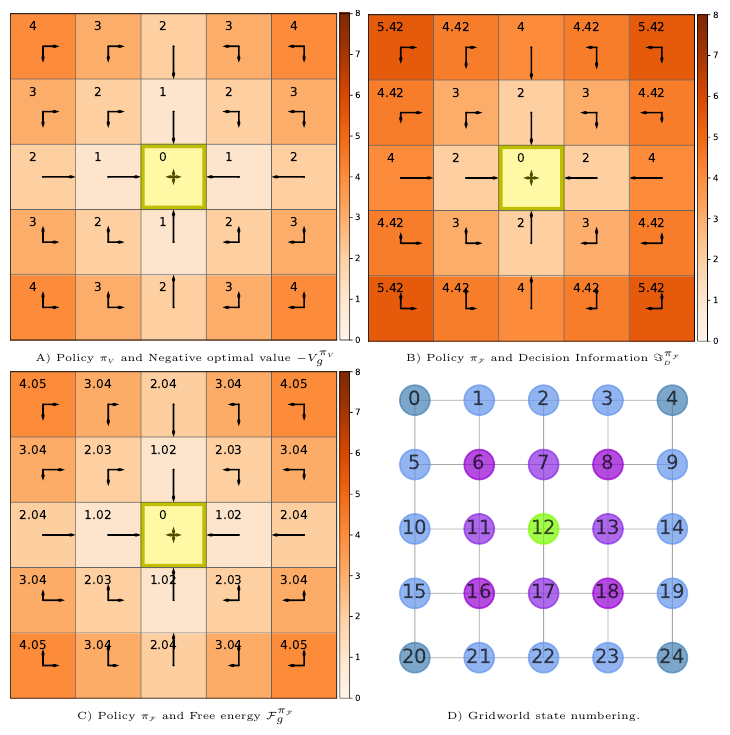}
\captionsetup{labelformat=empty} 
\caption{}
\label{fig1:decision-information}
\end{figure}

When following a value-optimal policy $\piv$, the agent selects a
sequence of actions with the aim of maximising the expected return, in
our case minimising the number of states visited (and actions taken)
on the way to the goal. \figurename \ref{fig1:decision-information}A
shows $\piv$ and the corresponding negative value function for a goal
in the centre $g=\#12$ as a heat map for a $5\times5$ gridworld with
Manhattan actions. Each square in the grid is a state, with goal
states denoted in yellow.    We thus interpret the
optimal value function $V^{\piv}_g$, which is traditionally written
$V^*$, as the negative distance between the agent's position $s$ and a
given goal state $g$.  Hence, we also denote the distance to $g$ from
$s$ by $-V^{\piv}_g(s)$.

\figurename ~\ref{fig1:decision-information}D shows the gridworld represented by a graph. The nodes in this and future graphs represent states which are numbered starting from zero to $\vert \S\vert$, and increasing first along consecutive columns and then row-wise. The colours of the nodes distinguish between different categories of states, e.g.\ having different locations with respect to the centre and whether or not they lie on the diagonal. The edges between the nodes indicate possible one-step transitions between states via actions.  The arrows in A--C represent the policy for the
agent acting in that square. The length of the arrows are proportional
to the value of the conditional probability $\pi(a|s)$, with the
action $a$ indicated by the arrow direction. \figurename \ref{fig1:decision-information}B and C will be
explained in upcoming sections \ref{decision-information} and \ref{constraining-information-processing} respectively. The numerical simulations used are detailed in section \ref{numerical-simulations}.

\clearpage
\marginpar{
\vspace{.7cm} 
\small
\color{captionColour} 
\textbf{Figure \ref{fig1:decision-information}. Value, Decision Information and Free energy Plots} in a $5 \times 5$ gridworld with cardinal (Manhattan) actions $\A:\{\shortarrow{0}, \shortarrow{2}, \shortarrow{4}, \shortarrow{6}\}$.  The goal $g = \#12$ is in the centre and is coloured yellow in the grid plots.  The arrow lengths are proportional to the conditional probability $\pi(a|s)$ in the indicated direction.  The relevant prior, i.e. the joint state and action distribution marginalised over all transient states, $\hat{p}(a;\pi)$ is shown in the yellow goal state. \textbf{A:} The policy displayed is the optimal value policy $\piv = \argmax_\pi V_g^\pi(s)$ for all $s\in\S$. The heatmap and annotations show the negative optimal value function $-V^\piv_g(s)$ for each state.  \textbf{B:} The policy presented is optimal with respect to free energy, i.e $\pif = \argmin_\pi \Fen_g^\pi (s;\beta)$ for all $s \in \S$. The heatmap and annotations show Decision Information $\ItgD^\pif(s)$ with $\beta$ = 100. \textbf{C:} The policy displayed is again $\pif$ with $\beta = 100$.  The heatmap and annotations show free energy $\Fen^\pif_g(s;\beta)$. \textbf{D:} Graph showing the numbering of states in the gridworld, the goal is coloured in green and the other colours indicate levels radiating from the centre.}

\subsection{The Value Function as a Quasimetric}\label{value-function-quasimetric}
We now discuss several concepts required for our geometric discussion.

A real valued function $d(x,y)$ is said to be a quasimetric for the space $\mathcal{X}$ if it satisfies the following conditions for all points $x,y,z$ in $\mathcal{X}$:
\begin{enumerate}[label={D\arabic*}]
	\item\!: \label{item:1} Nonnegativity:  $d(x,y) \geq 0$,
	\item\!: \label{item:2} Principle of Indiscernibles: $d(x,y)=0$ only if $x=y$,
	\item\!: \label{item:4} Triangle inequality: $d(x,y) + d(y,z) \geq d(x,z)$.  
\end{enumerate}

We begin by showing by simple arguments that the negative optimal value function is a quasimetric on the state space $\S$.

As the cumulative return is nonpositive $R_t \leq 0$, for the negative optimal value we have $-V^\piv_g(s)\geq0$, i.e.\ its value is nonnegative for all $s, g \in \S$, thus \ref{item:1} holds.

All the rewards from transient states are here chosen to be $-1$ i.e.\ $r(s'=\cdot, s=q, a=\cdot) = -1$ for all $q \in \{\S \setminus g\}$, apart from the transitions where the goal is the starting state in which case the reward is zero, i.e.\ $r(s'=g, s=g, a=\cdot) = 0$.  This means that $-V^\piv_g(s) = 0$ can only be obtained if $s$ is already the goal itself, 
thus \ref{item:2} holds.  

We can express the triangle inequality for the value function as:
$-V^{\piv}_{s'}(s) - V^{\piv}_g(s') \geq - V^{\piv}_g(s)$, for all $s,
s', g \in \S$, where $s'$ is some arbitrarily chosen intermediate
state. This can be easily proven by contradiction. In fact, let us
assume that one had $s,s',g$ with $-V^{\piv}_{s'}(s) - V^{\piv}_g(s')
< - V^{\piv}_g(s)$. By construction, $-V^{\piv}_{s'}(s)$ is the
shortest distance between $s$ and $s'$, and $- V^{\piv}_g(s')$ is the
shortest distance from $s'$ to $g$. This means that one could travel from $s$ to $s'$ and from $s'$ to $g$ with a total cost that is less than $- V^{\piv}_g(s)$, but that contradicts the choice of the latter as the shortest distance between $s$ and $g$. 

Therefore, \ref{item:4} holds.

Now, in a stochastic MDP the transition probabilities in a direction
may vary from those in the reverse direction. For example, actuators
may be noisier in one direction than in the other, e.g.\ due to
friction or ratcheting effects. Therefore, the value of an arbitrary
policy $V^\pi_g(s)$ is in general asymmetric with respect to $s$ and $g$, which is also true for its negative. This means that, while the negative optimal value function  fulfils the axioms of a quasimetric, it will not, in general, satisfy the symmetry axiom  required for it to act as a metric. 

\subsection{Information Processed in a Decision Sequence}

Up to now, we only  considered the raw cost of carrying out a policy from a starting to a goal state. However, in the spirit of resource-rational decision-making, we now make the transition to information measures  which consider a trade-off between performance and information, for a sequence of decisions. Such  measures were introduced and used in \citep{Tishby2011:information-to-go, Rubin2012,Larsson2017,Grau-Moya2019, Leibfried2019}. In each case, constraints on information processing are formulated using a generalisation of rate-distortion \citep[see chapter 10]{Cover2006}, permitting one to compute them via a Blahut-Arimoto-style algorithm \citep{Blahut1972, Arimoto1972}, alternating iterations of the constrained policy and an objective function (state-action value function $Q$, see section \ref{S-value-function} \eqref{S-eq:state-action-value-function}). Following \citep{Larsson2017}, we compute the policy which minimises free energy (see section \ref{constraining-information-processing} \eqname \ref{eq:free-energy}), for which we then compute Decision Information (see Algorithm \ref{S-alg:policy-free-energy} in \ref{S-computing-itg}). In the remainder of this section we will characterise these aforementioned measures in view of our objective to find a geometric representation. They typically differ according to the priors they use to which represent any previous knowledge concerning the expectations of state and action distributions in all possible future trajectories.

\textit{Information-to-go} \citep{Tishby2011:information-to-go} quantifies the information processed by the whole agent-environment system over all possible future state-action sequences.  It is the KL divergence between the joint probability of future trajectories conditioned on the current state and action, and the product of the marginal state and action distributions which corresponds to the prior expectation about the future. As such, it measures the information required to enact the trajectories from a state by the whole agent-environment assuming the first action has been taken. Its contributions can be decomposed into separate components, one for decision complexity and another for the response of the environment (see \ref{S-information-to-go}).  

Subsequent to this work on Information-to-go, \citeauthor{Rubin2012} \cite{Rubin2012} define \emph{InfoRL}, also referred to as ``control information'', as the relative entropy at state $s$ between the controller's policy and a fixed prior comprised of uniformly distributed actions, i.e. ``default plan'' (see \ref{S-InfoRL}). InfoRL is a maximum entropy cognitive cost, and it does not take into account the reduction in cognitive cost when, for example, the goal is either East or North of every state and thus only half the action space is at all relevant in every state.

\citet{Larsson2017} redresses this limitation in their definition of \textit{Discounted Information} which is aimed at optimising the use of computational resources. A single action is chosen to maximise rewards using a fixed probability distribution over states, thus, Discounted Information identifies an optimal prior action distribution which minimises the information cost across all states on average:  (see \ref{S-discounted-information}).  This is a discounted version of the decision complexity term of Information-to-go. The prior action distribution is here calculated by marginalising the policy over a fixed probability of states.  Discounted Information is the measure most similar to the measure we use below (see \eqname \refeq{eq:itg-d-bellman}), with the main difference being that it uses a fixed probability distribution over states to calculate the prior.

Mutual Information Regularisation (MIR) \citep{Grau-Moya2019} (see \ref{S-mutual-information-regularisation}) extends the information processed in a single decision, referred to as ``one-step entropy regularisation'' to a sequence of decisions by defining a discounted value function which incorporates the mutual information between states and actions.  The optimal policy is the marginal distribution over actions under the discounted stationary distribution over states  \citep[see equations (4) and (8)]{Leibfried2019}. 

\subsection{Decision Information}\label{decision-information}

As the remit of this work is to investigate how the space of goals
transforms under information processing constraints, we would like to
use an information measure which, through our choice of prior, takes
into account the actual behaviour induced by the agent's actions, or,
more precisely, policy.   Additionally, we wish to disregard the information cost of deviations from the marginal actions in states that will not be visited by the agent.  
At the same time we require a measure of the expected cognitive cost
where actions in the past and future carry the same weighting, and we
therefore obviate discounting.  Information-to-go is conditioned on
the first state and action pair and therefore, it does not include the
first decision in its cognitive cost.  In addition it also includes
the information processed by the environment to determine the
successor state resulting from an action. Optimising both the decision
term and the environmental response term results in optimal policies
which prefer short trajectories and retain a substantial element of
goal-directed behaviour even when significant constraints are applied
to information processing. 

We thus formalise the information processed in a sequence of decisions via
a quantity derived from Information-to-go where we drop the
conditioning on the current action and condition only on the current
state.  Furthermore, we extract exclusively the component which impacts
the decision complexity as we are only interested in the information
cost for the decision-maker. This also suppresses the
goal-directedness induced by the decision complexity term. Any such
goal-directedness will therefore be entirely captured by the value
function. Formally, we define \textit{Decision
  Information} for a policy $\pi$ as the decision cost of an agent
following $\pi$ for future trajectories starting in $s$, as per
\eqname\eqref{eq:itg-d-bellman}, with more details available in
\ref{S-decision-information}. We assume that the prior
$\hat{p}(a;\pi)$, see \eqname\eqref{eq:marginal-action-distribution},
is the joint distribution of states and actions marginalised over a
\textit{``live'' state distribution} (explained below), which is
denoted by $\hat{p}(s;\pi)$, as shown in \eqname
\eqref{eq:marginal-action-distribution}. We use the notation
$p(\cdot; \pi)$ to identify $\pi$ as a parameter of the distribution.

\begin{equation}
	\ItgD^{\pi}(s) \defeq \E_{\pi(a|s)p(s'|s, a)}\left[\log{\frac{\pi(a|s)}{\hat{p}(a; \pi)}} + \ItgD^{\pi}(s')\right]. \label{eq:itg-d-bellman}
\end{equation}
\begin{equation}
    \hat{p}(a; \pi) = \sum_{s \in \mathcal{S}}\pi(a|s)\hat{p}(s; \pi). \label{eq:marginal-action-distribution}
\end{equation}

Given a policy, the ``live'' or ``visitation'' state distribution (see
\ref{S-live-state-distribution}) is a stationary state distribution
modified for an absorbing Markov chain where the probability mass in
absorbing or terminal states is suppressed to model the typical state
of the system while not yet in an absorbing state \citep[see section
11.2]{Grinstead2006}. Without that, an unmodified stationary state
distribution will result in the bulk of the probability mass being located
in the goal state after a sufficiently long time, with the probability
of all other states being visited tending to zero.
The live distribution on the state $s$ expresses the probability of
finding the agent at a particular non-goal state at some random time
of carrying out its policy $\pi$, if it originally started its
trajectory uniformly somewhere in the state space, at some
$s \in \S \setminus g$. It measures how probable it is to find an
agent at an intermediate transient state en route to the goal and
thereby represents the probability of the agent having to take a
decision in that particular state while its policy is ``live''. We
assume that the agent's location follows a live state distribution to make Decision Information coherent with the overall a priori probability of
the agent being in the state in question while still underway to a
goal state.

The prior action distribution $\hat{p}(a;\pi)$ encodes all information known a priori about the action process where we assume that the joint action distribution can be factorised by policy-consistent distributions according to $\Pr(A_t=a_t, \dots, A_T=a_T) = \hat{p}(a_{t})\cdot \hat{p}(a_{t+1}) \cdots \hat{p}(a_{T})$. 
We also assume that for the joint state distribution, 
$\Pr(S_{t+1}=s_{t+1}, \dots,S_T=s_T) = \hat{p}(s_{t+1})\cdot
\hat{p}(s_{t+2}) \cdots \hat{p}(s_{T})$.  As in  \citep[section
6.2]{Tishby2011:information-to-go} 
we assume that the probabilities $\hat{p}(s_{t+1}), \hat{p}(s_{t+2}),
\dots$ are time-homogeneous.  Therefore, the action prior is also
time-independent as it is calculated using the state prior as per
\eqname \eqref{eq:marginal-action-distribution}.  In the
aforementioned literature, the prior state distribution is not
parameterised by the policy $\pi$, but rather distributed according to
a uniform distribution \citep{Tishby2011:information-to-go, Rubin2012,
  Larsson2017}. Using uniform distributions as the state prior
has the problem that the distribution fails to take into account that
some states may be unlikely to be visited, except in the case of
rarely travelled trajectories, while others (e.g.\ close to a goal)
will concentrate far more probability mass. Furthermore, in contrast
to a stationary distribution, a live distribution excludes all
recurrent states (here, goal states) and reflects the distribution
only of states visited by the agent while the latter still needs to
make decisions. It is important to note that, while at any given time
step this prior does not depend on the states traversed in the
previous time steps of a given trajectory,  the live distribution will
still reflect the overall visitation frequency of states induced by the current overall policy. Furthermore, Decision Information raises more costs in states in which decisions differ more from the typical behaviour(s) in the transient states visited by the policy.

Figure~\ref{fig1:decision-information}B shows the Decision Information
values for each state given a policy $\pif$ which is optimal with
respect to free energy which is defined below. We use the goal state specifically
to represent the action marginal $\hat{p}(a;\pi)$ rather than the
actual action selection because, regardless of actions selected in
this state, the agent remains in the goal and, per definition, this
does not contribute to the Decision Information, thus displaying any actual action selection
in the goal would therefore not be of interest. This enables a convenient comparison with the denominator in \eqname\eqref{eq:itg-d-bellman} when interpreting Decision Information.

\subsection{Constraining Information Processing}\label{constraining-information-processing}
The traditional MDP framework assumes the agent is able to access and process all information necessary for an optimal decision. However, when information processing resources are scarce, we want the agent to prioritise the processing of essential information only. This can be achieved, for instance, by compressing state information as much as possible without compromising performance \citep{Polani2006}. Information-theoretically, one can consider the sequence of states in the agent's trajectory as input and the respective actions taken as output; in analogy to rate-distortion theory \citep{Cover2006}, one then seeks the lowest information rate required to reach a certain value. 

Formally, we seek a solution to the following constrained optimisation problem, with the desired information rate $\tilde{\ItgD}(s)$ determining the trade-off between information processing and performance: \begin{equation}
	\max_{\pi(a|s)}V ^{\pi}(s) \text{ s.t. } \ItgD^{\pi}(s) = \tilde{\ItgD}(s). \label{eq:itg-constrained}
\end{equation}
For optimisation, we write this as an unconstrained Lagrangian (\eqname \ref{eq:itg-lagrangian}), with the Lagrangian multiplier $\beta$ for the information rate constraint $\tilde{\ItgD}$ and $\lambda$ for the normalisation of the policy as shown in \eqname\eqref{eq:itg-lagrangian}.  Here we consider overall resource constraints for the sequence and not per-step bandwidth constraints.
\begin{equation}
	\mathcal{L}^{\pi}(s;\beta, \lambda) \defeq \frac{1}{\beta}\ItgD^{\pi}(s) - V^{\pi}(s) + \lambda\left(1-\sum_{a}\pi(a|s)\right). \label{eq:itg-lagrangian}
\end{equation}

Note that, $\lambda$ is chosen in such a way that, on optimisation
of the Lagrangian, the associated bracketed term disappears, which
reflects the fact that the policy defines a probability distribution
over the actions which is normalised. Thus, the Lagrangian reduces
to the following terms, which we call the \emph{free energy} of our problem, as seen in \eqname\eqref{eq:free-energy}. Free energy is the trade-off between expected utility and Decision information, i.e. the cost of the information processing \eqname \eqref{eq:itg-d-bellman} required to execute a behaviour policy $\pi$ \citep{Ortega2015, Larsson2017}.    In our cost-only MDPs $\ItgD^{\pi} \geq 0$ and $V^{\pi} \leq 0$, thus their free energy is always nonnegative.  The trade-off parameter $\beta$ will be strictly non-zero.  
\begin{equation}
	\Fen^{\pi}(s;\beta) \defeq \frac{1}{\beta}\ItgD^{\pi}(s) - V^{\pi}(s).\label{eq:free-energy}
\end{equation}

In principle one might choose a different Decision Information threshold $\tilde{\ItgD}(s)$ for each state as per \eqname \eqref{eq:itg-constrained}. However, there are both conceptual as well as practical difficulties in choosing per-state thresholds consistently \citep{CatenacciVolpi2020:goal-directed-empowerment}; we therefore instead follow their approach and choose a single $\beta$ for the whole system. This corresponds to a different threshold for every state $s$, 
which is not only computationally convenient, but also turns out to give more immediately useful results than other schemes that were explored for per-state threshold selection (at this stage, no theoretical justification for this observation is given). With this setting, a double iteration combining the Bellman equation with the Blahut-Arimoto algorithm \citep{Blahut1972} computes the free energy, see \citep{Larsson2017} and the algorithm \ref{S-alg:policy-free-energy} in \ref{S-minimising-free-energy}. Minimising the Lagrangian when $\beta$ is very small yields a policy $\pif$ where Decision Information becomes the dominant term \eqref{eq:itg-lagrangian} and the agent aims to minimise information processing at the expense of increasing distance cost.

From value \eqref{eq:value-bellman} and Decision Information
\eqname\eqref{eq:itg-d-bellman}, free energy inherits the form of a
Bellman equation shown in \eqname\eqref{eq:free-energy-bellman} where
the ``new'' value for the single recursive step (e.g.\ reward addition
in the case of an MDP) is combined with the sum of future rewards.
Decision Information and the value function are both expected values
taken over all future trajectories starting from the current state $s$
and following the policy $\pi$ thereafter. Free energy, which is
nonnegative, is thus the expected value of a weighted combination of
the information processed and rewards accumulated over all future
trajectories, see \eqname\eqref{eq:free-energy}. The optimal free
energy $\Fen^\pif(s; \beta)$ is taken at the fixed point of this
equation using the corresponding free energy optimal policy $\pif$,
\citep{Tishby2011:information-to-go,Rubin2012,Larsson2017}. This
policy $\pif$ is the policy which maximises performance given a
constraint on information processing. For details on finding $\pif$
see \ref{S-computing-itg} in the supplementary material. Referring
again to Fig. \ref{fig1:decision-information}, plot C shows free
energy values for a $\beta$ value of 100, which correspond to the
Decision Information values plotted in \figurename 
\ref{fig1:decision-information}B. Here, the trade-off value is
weighted strongly towards value performance, thus the behaviour
strategy is close to the value-optimal behaviour shown in  \figurename 
\ref{fig1:decision-information}A. The free energy
calculated over policy $\pif$ monotonically decreases from the current
state to the goal (in expectation).

\begin{align}
\Fen^{\pi}(s;\beta) 
&=\E_{\pi(a|s)p(s'|s, a)}\biggl[ \frac{1}{\beta} \log \frac{\pi(a|s)}{\hat{p}(a)} - r(s', s, a)  + \Fen^{\pi}(s') \biggr].\label{eq:free-energy-bellman}
\end{align}

We have disambiguated the notation regarding optimal policies,
conventionally written as $\pi^*$, by removing the asterisk and
instead using a subscript, to distinguish $\piv$ being optimal with
respect to the value function from $\pif$ being optimal with respect
to free energy. Following an optimal policy from state $s$ to goal
state $g$, the agent is able to follow a policy, with an 
expected free energy at a given level of performance denoted by $\Fen^\pif_g(s)$. For
the limit case of Decision Information approaching zero, the policy
$\pif$ converges towards the marginalised action distribution in all
live states. As this is the same in all states, i.e.\ state independent, it is an open-loop policy.  One can consider this to be the discrete-world probabilistic generalisation
of moving in a continuous space along a fixed direction. It additionally tries optimising the value achievable
for this open-loop policy.

\subsection{Numerical Simulations}\label{numerical-simulations}
We conducted simulations of a navigational task in a 2D gridworld
modelled as a cost-only MDP with a discrete and finite state space.
While here considering  deterministic transitions, the framework
generalises smoothly to the non-deterministic case. Details of the algorithm and the code repository\footnote{Data and relevant code for this research work are stored in \url{https://gitlab.com/uh-adapsys/cognitive-geometry/} and have been archived within the Zenodo repository: \url{https://zenodo.org/record/7273868}.}   are included in \ref{S-code-repository} in the supplementary material.  Starting from an initial state,  at each time step the agent selects an action representing a move in the given neighbourhood: in Manhattan, with action set $\A : \{\shortarrow{0}, \shortarrow{2}, \shortarrow{4}, \shortarrow{6} \}$, in  Moore, with  action set $\A : \{ \shortarrow{0},  \shortarrow{1}, \shortarrow{2},  \shortarrow{3}, \shortarrow{4},  \shortarrow{5}, \shortarrow{6}  \shortarrow{7} \}$.  

In our model, if the agent bumps into a wall, the agent does not move
but still receives a reward of $-1$, there is thus no additional
deterrent for an agent to bump into boundaries, apart from loss of
time. The edges and corners are buffering incorrect actions to some
extent. Practically, this facilitates successful near open-loop
policies at low $\beta$ values where incorrect actions can be
frequent. 

For visualisation, we used a non-linear mapping algorithm, Multidimensional Scaling (MDS) \citep{Kruskal1978}, which, given the symmetrical pairwise distances between points in a set,  projects each point onto an $N$-dimensional space such that the distances between objects are preserved as much as possible.  We use the Scikit-learn MDS algorithm \citep{scikit-learn} to qualitatively assess the distance-related aspects of geometry imposed by the trade-off between value (grid distance) and information.  To model cognitive distances, we compute free energies $\Fen^\pif_{s_j}(s_i)$ between all combinations of pairs of states, $s_i , s_j \in \S$, creating an $n\times n$ pair-wise matrix $D$, with elements $d_{ij} \defeq \Fen^\pif_{s_j}(s_i)$.  In order to visualise these distances in 2D/3D Euclidean space, $D$ is symmetricised into $D^{\mathrm{sym}}$ by taking the average of the reciprocal trajectories, i.e.\  $d^{\mathrm{sym}}_{ij} \defeq \left(d_{ij}+d_{ji}\right)/{2}$, hence 
\begin{equation}
	D^{\mathrm{sym}} \defeq \left(D + D^{\mathop{\intercal}}\right)/{2}. \label{eq:D-sym}
\end{equation} 

Figures \ref{figS2:frees-man-b-100} and \ref{figS3:frees-man-b-0.1} in the supplementary material show the free energy values and optimal policies for every combination of $(s,g)$ for $\beta=1\times10^7$ and $\beta = 0.1$ respectively.  Figures \ref{figS4:frees-asym-prop-man-b-0.1} and \ref{figS5:frees-asym-prop-man-b-0.01} in the supplementary material show the discrepancy between the symmetrised and non-symmetrised free energy values for $\beta$ values of 0.1 and 0.01 respectively.  Asymmetry comes into play particularly in the regions between the centre and the corner states.  

\section{Results}
We now elucidate the nature of Decision Information, proceeding to
demonstrate the trade-off between performance and cognitive burden,
and then present visualisations of distortions that emerge under
optimal policies, showing how constraints on information processing
impact the geometry of the gridworld. We later explain that we
interpret structure as a collection of optimal routes or ``desics'':
either geodesics which are optimal in terms of the value function (see
section \ref{value-geodesics}) or ``infodesics'' which are optimal
with respect to free energy and thus take information processing into
account (see section \ref{infodesics-defined}). In what follows, the
optimal policy $\pif$ is that which minimises the free energy, and
$\piv$ the special case when we maximise the reward (minimise the
number of steps to reach the goal), i.e.\ the limit case obtained by
setting $\beta \to \infty$ in the free energy.

\subsection{Decision Information}

\begin{figure}[!ht]
\includegraphics[width=\textwidth]{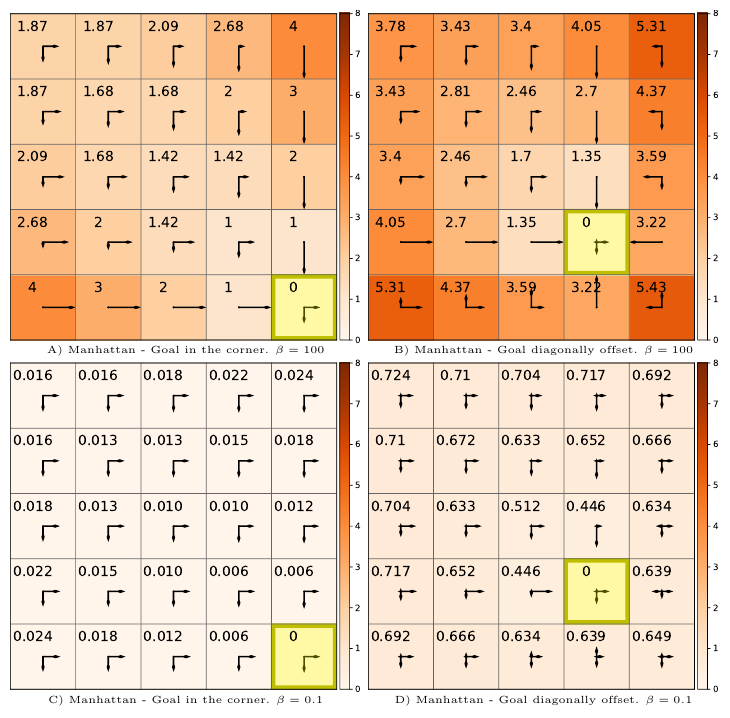}
\captionsetup{labelformat=empty} 
\caption{}
\label{fig2:Itg-policy-grid-man}
\end{figure}

Decision Information computes the information processed for the
\emph{whole} sequence of decisions leading the agent to the goal. Thus
it depends on the length of the sequence.  This accumulation of cost is readily seen in \figurename \ref{fig2:Itg-policy-grid-man}A on the trajectory moving outwards from the goal along the lower edge, where each action adds one bit to the information processing cost and longer routes thus become informationally more expensive. 

From \eqname \eqref{eq:itg-d-bellman}, we see that Decision Information increases when the alignment between the policy in a particular state and the marginalised action distribution $\hat{p}(a;\pi)$ decreases. \figurename \ref{fig2:Itg-policy-grid-man}B shows Decision Information for a goal state in between the lower right corner and the middle of the grid, $g=\#18$.  The reader is advised to refer to \figurename \ref{fig1:decision-information}D for state numbering.  Starting from below or right of the goal, the agent is required to select actions which oppose the prevailing actions in the marginalised action distribution, causing a noticeable increase of Decision Information in these squares.  For example, consider the Decision Information in the states adjacent to the goal: $\ItgDpif_{g=\#18}(23)=3.22$ bits directly below the goal, in contrast to $\ItgDpif_{g=\#18}(13)=1.35$ bits directly above the goal.  

Decision Information is asymmetric in general as it is a function of
the policy and thus dependent on the information processing required
for a particular goal, for example, moving from the corner $s=\#24$ to
the centre $g=\#12$  carries a decision cost of $
\ItgDpifi{1}_{g=\#12}(24) = 5.42$ bits (\figurename
\ref{fig1:decision-information}B), in comparison to  $
\ItgDpifi{2}_{g=\#24}(12) = 1.42$ bits from the centre $s=\#12$ to the
corner $g=\#24$ (\figurename \ref{fig2:Itg-policy-grid-man}A).  The
cost of accurately stopping exactly in the centre is higher than
stopping at the edge where some of the cost is borne by the embodiment
as the agent is blocked by the wall from moving beyond the goal in the corner.

\marginpar{
\vspace{.7cm} 
\small
\color{captionColour} 
\textbf{Figure \ref{fig2:Itg-policy-grid-man}. Decision Information heat map}
in a $5 \times 5$ gridworld with $\A:\{\shortarrow{0}, \shortarrow{2}, \shortarrow{4}, \shortarrow{6}\}$.   Annotated values show $\ItgDpif(s)$ in bits.  The policy presented is optimal with respect to free energy, i.e $\pif =  \argmin_\pi \Fen^\pi (s;\beta)$.  The arrow lengths are proportional to the conditional probability $\pi(a|s)$ in the indicated direction, for convenience the prior $\hat{p}(a)$ is shown instead of the policy in the yellow goal state. \textbf{Reward-maximising behaviour ($\beta = 100$):} In \textbf{A:} the goal is in the corner and in \textbf{B:} the goal is on the diagonal between the corner and the middle. \textbf{Behaviour where information processing is constrained ($\beta=0.1$)}: \textbf{C:} the goal is in the corner and in \textbf{D:} it is diagonally adjacent to the corner. }

\subsection{Visualising Cognitive Geometry}\label{visualising-cognitive-geometry}

\begin{figure}[!ht]
\includegraphics[width=\textwidth]{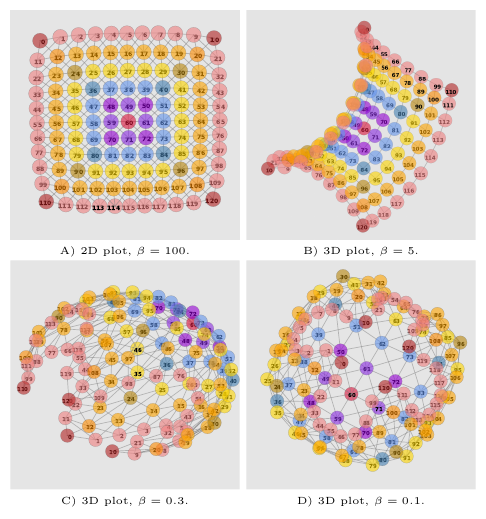}
\captionsetup{labelformat=empty} 
\caption{}
\label{fig3:mds-moo-11-11}
\end{figure}

Cognitive geometry shapes the way agents represent the distance
between states of the environment when information costs are taken
into account. By way of introduction to our discussion on generalising
geodesics to rationally bounded agents (see section
\ref{generalising-geodesics}) we begin by presenting visualisations of
our cost-only gridworlds subjected to information constraints over a
range of $\beta$ values. \figurename \ref{fig3:mds-moo-11-11} shows
two- and three-dimensional MDS embeddings for the symmetrised free energy
pairwise distances \eqref{eq:D-sym} in an $11 \times 11$ gridworld
with a Moore neighbourhood for different $\beta$ values. In essence,
free energy trades off distance travelled against the specificity of
the policy as regards the state. The balance between these two factors
is determined by the value of $\beta$. So when states are close
together in terms of free energy then one needs to interpret this in
light of the trade-off parameter $\beta$. If an agent behaves
optimally with respect to value alone (approximated by $\beta = 100$),
then the free energy reduces essentially to the distance travelled
between the states, reconstituting simply the original Moore geometry,
as indicated in \figurename \ref{fig3:mds-moo-11-11}A, where the
underlying grid is prominently discernible. The difference between the
original and the symmetrised free energies also varies with $\beta$ as
the informational component is dependent on the policy. The
symmetrised free energy adjacency matrix is merely used to provide a
suitable approximation for its visualisation. 
The discrepancies between real and symmetrised free energies are
shown in section \ref{S-symmetrised-free-energy} of the
supplementary material in \figurename
\ref{figS4:frees-asym-prop-man-b-0.1} and
\ref{figS5:frees-asym-prop-man-b-0.01}. 

\marginpar{
\vspace{.7cm} 
\small
\color{captionColour} 
\textbf{\figurename \ref{fig3:mds-moo-11-11}. MDS visualisation of the cognitive geometry} induced by free energy in a Moore $11 \times 11$ gridworld. As $\beta$ decreases, the corners migrate towards each other and the topology morphs from a flat grid to a mesh wrapped over a ball. \textbf{A} At $\beta = 100$, the free energies approach the optimal value function $V^*$.  With the reduction of $\beta$, \textbf{B} $\beta = 5$, \textbf{C} $\beta = 0.3$ and \textbf{D} $\beta = 0.1$.
 
As a result of the constrained information processing, the policy results in similar actions taken in multiple states, i.e.\  the agent moves in consistent directions across the grid.  The routes to the corners and along the edges are more informationally efficient, with the effect that corners migrate towards each other, despite being farthest away from each other in the original grid distance. }

In our gridworld configuration, an agent can take repeated actions to move diagonally and also get additional guidance along the edges, as bumping into them incurs only a minor time penalty. Thus the wall acts as a guide or ``funnel''  buffering some wrong actions on the way to the corner goal.  In contrast, to reach a middle goal, any wrong action is likely to move the agent away from the goal. Thus much more accurate control is required to reach precisely the middle goal compared to drifting into a corner.  The latter decision strategy can be likened to \textit{ecological rationality}, where an agent makes use of simple heuristics which specifically exploit the structure of the environment \citep{Lieder2020:resource-rational-cognition}. 

Note that, if walking into walls would be
penalised, the result would be substantially different. In this case,
free energy distances would increase rather than decrease by closeness
to walls and corners would generally tend to be further away (``pushed
out'') in the free energy metric rather than closer together (``pulled
in'') as they are in our experiments.  We also emphasise that the insertion of any obstacles in the
world will create various, not necessarily intuitive distortions which
would reflect which task groups would now become easier or more
difficult to do together. 

Since the informational cost of executing an optimal policy thus
depends drastically on the nature of the goal, our distance, once
taking this informational cost into account, will significantly
distort the geometry of the gridworld as compared to a naive spatial
map. When information processing is restricted via progressively lower
$\beta$ values, it becomes the cost of information processing that
that dominates the
optimisation compared to the extra time cost for bumping into the wall, and policies which utilise the guiding property of the edges are increasingly preferred.  Thus, states on the edges, even more so in the case of the corners, effectively become closer in terms of free energy ``distances'', as if the spatial geometry were wrapped around a ball with the corners drawn together, see \figurename \ref{fig3:mds-moo-11-11} C \& D.

\section{Value geodesics}\label{value-geodesics}
Armed with the observation that, given the reward function as defined
previously for cost-only MDPs, the value function forms a quasimetric
over the state space  (section \ref{value-function-quasimetric}), we now
define a \textit{value geodesic} from $s$ to $g$ to be a sequence of
arbitrary finite length $\tau_{s \rightarrow g} = \langle s_0=s, s_1, s_2, \dots, s_N=g\rangle$, $s_i\in\S$ such that all elements of the sequence turn a generalisation of the triangle inequality into an equality, i.e.\ \begin{equation} 
	V^\piv_{g}(s) = \sum_{i=0}^{N-1} V^\piv_{s_{i+1}}(s_{i})~ \text{with } s_i \in \tau_{s \rightarrow g}.\label{eq:geodesic}
\end{equation}

Such a value geodesic does not in general consist of a contiguous
sequence of states, however, in the case of cost-only MDPs, they are. Furthermore, every state $s'$ on the value geodesic lies on a shortest path from the initial state $s$ to that intermediate state $s'$. So, not only is the path from $s'$ to the goal $g$ optimal (this is a consequence of the Bellman property), but also the path from $s$ to the intermediate states $s'$. This has an important consequence: each state on this sequence can be considered a potential alternative goal that can be reached optimally from $s$ while en route to $g$. It means that benefits from reaching these intermediate goals can be ``scooped up'' on the way to the main goal. Within this sequence, the value is strictly monotonically increasing, i.e.\ $ V_g(s_n) < V_g(s_m)$ for $n<m$. 

For cost-only MDPs, the policy in the value functions of different
subsegments of a value geodesic can be taken to be the same. This is
because being the (negative of the) value, a quasimetric, and given
that the geodesic is the shortest path from $s$ to $g$ which can be
traversed using the optimal policy $\piv$, then also every subsegment
from $s_i$ to $s_j$ of this path is also the shortest path from $s_i$
to $s_j$ that can be traversed using the same policy $\piv$. As
opposed to this, in the next section we will show for infodesics,
which take into account the cost of information processing, in general
each subsegment has a different optimal policy.

\section{Infodesics}\label{infodesics-defined}

Motivated by the concept of value geodesics, we propose an analogous
concept in our space where distances are determined partially or
completely by an information-theoretic component. We term this concept
\emph{infodesics}.  We will first consider \emph{pure infodesics} as
sequences of states that can be visited by the agent using no state
information at all, i.e.\ with action selection being independent of
state. The intention is that, as with traditional geodesics, following
infodesics one obtains all interim states optimally, at no extra
decision cost after choosing the initial direction (i.e.\ optimal
open-loop policy in our perspective).  In a discrete setup, we do not have intermediate directions, and since we
are not using memory to enforce particular movement patterns (e.g.\
moving diagonally in a Manhattan world by alternating taking one step
north followed by one step east, for instance), a pure geodesic would
be most closely represented by a fixed action distribution over
all states. Now, Decision Information measures the deviation from the
action marginal, i.e.\ from a state-independent policy; thus, its minimisation
corresponds to the attempt to minimise the cumulative expenditure in
information processing due to deviating from that marginal. 
Hence, Decision Information can be considered a quantity that, when minimised, tries to approximate following the same direction in the continuum with as little cognitive correction as possible.

We now define a \textit{cost infodesic} as a value geodesic where the
value function $V$ has been replaced by some cost  function, in our
case, the free energy function $\Fen$ (with the sign suitably
reversed).  In such a cost infodesic, action selection is not in
general restricted to being state independent.  A cost infodesic in a
discrete MDP is thus an arbitrarily, but finitely long sequence $\psi_{s\rightarrow g}= \langle s_0=s, s_1, \dots, s_N=g\rangle$, $s_i\in\S$ such that, for that sequence, all elements of the sequence turn a generalisation of the triangle inequality for free energy into an equality \eqref{eq:infodesic}. Formally, we demand for a cost infodesic that:
\begin{equation}
	\Fen^\pif_g(s) = \sum_{i=0} ^ {N-1} \Fen^{\pifi{i}}_{s_{i+1}}(s_{i}) ~ \text{with } s_i \in \psi_{s \rightarrow g} \text{ by abuse of notation}. \label{eq:infodesic}
\end{equation}
The value of $\Fen^\pif_g(s_i)$ for such a cost infodesic is strictly monotonically decreasing as shown below, $$\Fen^\pifi{n}_g(s_n) > \Fen^\pifi{m} _g(s_m) \quad \text{for all } n<m.$$  

Note that we do not require that the cost infodesic is contiguous
(i.e.\ consisting only of adjacent states) throughout the trajectory,
as is the case for cost-only MDPs in deterministic value geodesics;
also, nontrivial infodesics (i.e.\  infodesics containing more than start and goal state) do not need to exist, either. They therefore typically constitute a strongly impoverished collection of trajectory fragments compared to traditional deterministic geodesics.

These cost infodesics, with free energy as a distance measure, constitute our objects of interest.  The policy $\pif$
encodes a single behaviour strategy for an agent to travel from $s \to
g$ resulting in a route optimal in terms of free energy.  An infodesic
from $s \to g$ can be split into subsegments (which we will index by
$i$), which have an equivalent combined free energy cost.  If these
subsegments have different information costs, then, the optimal
policies for these subsegments are in general not equivalent and are thus distinguished using a superscript as in $\pifi{i}$.  

When following the policy that minimises $\Fen_g^{\pif}(s)$, given the
stochasticity of the policy, the agent is not guaranteed to pass
through any particular intermediate infodesics states between $s$ and
$g$ before reaching the goal.  In order to assure that the agent
visits these intermediate states, subgoals need to be added to the
navigation task.  Optimality on the subpaths $s_i$ to $s_{i+1}$ can be
achieved by allowing the agent to use multiple policies that minimise
the free energies $\Fen_{s_{i+1}}^{\pifi{i}}(s_i)$. Hence, if, for
instance, in the infodesics ${\langle s, s', g\rangle}$ we make $s'$ a
subgoal (i.e., the trajectory of the agent is constrained to pass
though $s'$), the trajectory from $s$ to $g$ can be formed by
concatenating the sub-trajectory from $s$ to $s'$ with the
sub-trajectory from $s'$ to $g$.    Interestingly, when the triangle
equality for free energy is thus enforced as a geodesics criterion, this concept of \textit{policy switching} and a collection of corresponding informationally salient midpoints emerges via the segmentation of the agent's behavior.  

We interpret geometry as
essentially the whole collection of ``desics'' under consideration,
i.e.\ directed trajectories which connect their points in a (locally)
optimal manner, according to the chosen measure. 
When one operates with a quasimetric in the background,  one obtains a
traditional collection of geodesics,  which together with their
starting and goal states show which goals can optimally be scooped up
en route.  In our picture this collection then constitutes  the
geometry of the given space. Now, however, we derive the infodesics
from the free energy which we later show is not a quasimetric  in general. Thus, several of the guaranteed properties of geodesics are lost.  Furthermore, this collection of infodesics is much sparser, and the infodesics themselves can be ``deficient'' compared to traditional geodesics in the sense that neither are the infodesics necessarily contiguous, nor even nontrivial. Nevertheless, the geometry still represents the collection of the various interrelations, even if now with weaker guarantees.

To regain some of the richness of structure of the traditional geodesics, we will therefore compensate for above limitations by treating the free energy as a ``deficient'' quasimetric and permit some relaxation of constraints.

To explore the analogy with the set of traditional geodesics, we
reiterate the idea: a geodesic is characterised by subsequences for
which the concatenation of each section turns the triangle inequality
into an equality\footnote{By assuming that the full task trajectory
  from $s$ to $g$ is a cheapest or shortest route, we ignore without
  loss of generality any technical complications from a possible
  wraparound of the trajectory.} For the infodesics, we proceed in the
same way, but since they are no longer guaranteed to be contiguous, or
even nontrivial at all, in our experiments, we instead perform an
exhaustive search over sequences formed by every non-repeating
combination of sequences of 3, 4 or 5 states. As in the geodesic case,
the states in this sequence are taken to be interim goals; free energy
and respective optimal policies were calculated for each subtrajectory
within the sequence. It is important to reiterate that in the context
of interim goals, the agent might follow a different optimal policy
$\pifi{i}$ for each separate portion of a trajectory (as is also
possible in the case for traditional geodesics split into segments).
Since information processing cost plays a major role, this observation
will turn out to be significant, not only because policies vary in
terms of their informational processing costs, but because switching
policies introduces additional cognitive costs which do not have an
analog in traditional geodesics.

In our setup, strict infodesics may be trivial, i.e.\ may end up not containing any intermediate states.  We therefore relaxed the generalised triangle inequality requirement for the infodesic and defined an $\varepsilon$\textit{-infodesic} such that the normalised difference between the sum of the free energy for the trajectory with interim goals and the total, single-goal free energy $\Fen^\pif_{s_T}(s_0)$, is within the range of $\varepsilon$, with $0 <  \varepsilon   \ll 1$ via \eqref{eq:infodesic-relaxed}. As we will show later, this normalised difference, however, can become negative, as is the case, discussed below, when, in the absence of considering the costs to switch between different policies,  it becomes advantageous to use multiple policies.
\begin{equation}
	-\varepsilon < \frac{[\sum_{i=0}^{T-1}\Fen^{\pifi{i}}_{s_{i+1}}(s_i)]-\Fen^\pif_{s_T}(s_0)}{\Fen^\pif_{s_T}(s_0)} < \varepsilon. \label{eq:infodesic-relaxed}
\end{equation}

\subsection{Examples of \texorpdfstring{$\varepsilon$} --Infodesics on Cost-only MDPs} 
\begin{figure}[!ht]
\includegraphics[width=133mm, trim= 5 0 0 0, clip]{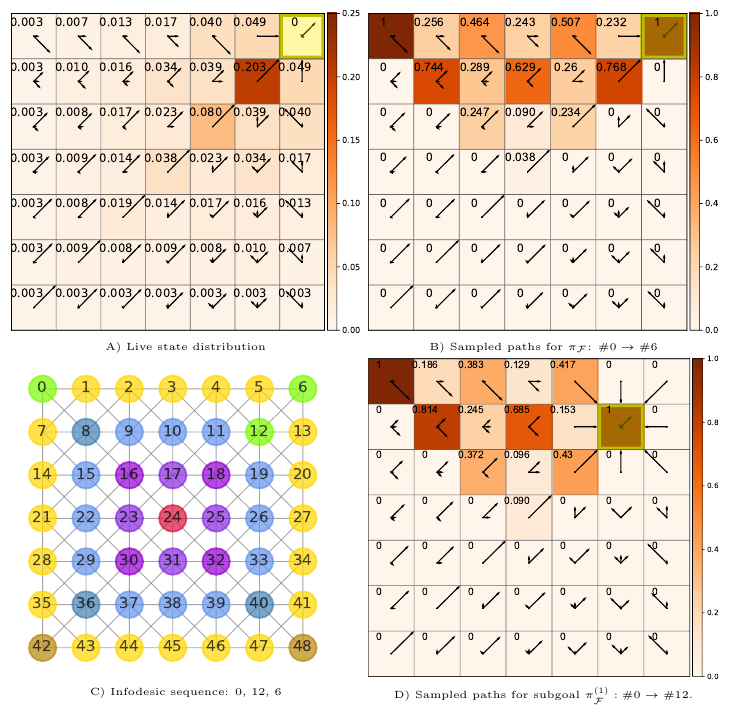}
\captionsetup{labelformat=empty}
\caption{}
\label{fig4:infodesic-moo-12-6-b-100}
\end{figure}

\begin{figure}[!ht]
	\begin{adjustwidth}{-2.0in}{0in}
		\includegraphics[width=\linewidth, trim= 5 5 7 0, clip ]{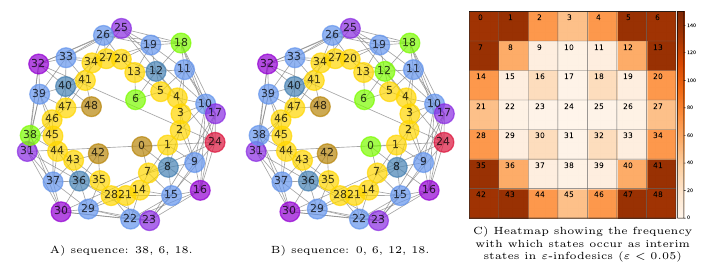}
		\captionsetup{labelformat=empty} 
		\caption{} 
		\label{fig5:infodesic-moo-6-18}
	\end{adjustwidth}
\end{figure}

\begin{figure}[!ht]
\includegraphics[width=133mm]{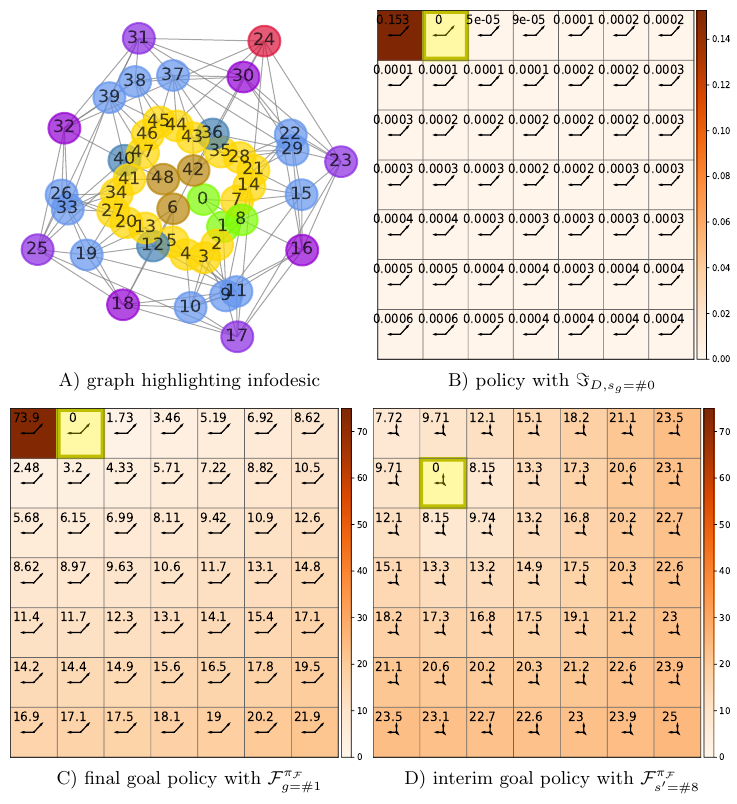}
\captionsetup{labelformat=empty}
\caption{}
\label{fig6:infodesic-moo-0-8-1}
\end{figure}

We study cost infodesics in a 7 $\times$ 7 Moore gridworld with three
sets of examples in order of decreasing values of $\beta$:
reward-maximising behaviour, at $\beta=100$; limited information
processing, at $\beta = 0.07$; and near-minimal information
processing, at $\beta = 0.01$.  Figures
\ref{fig4:infodesic-moo-12-6-b-100}, \ref{fig5:infodesic-moo-6-18} and
\ref{fig6:infodesic-moo-0-8-1} present these infodesics as sequences
of not necessarily contiguous interim states which are highlighted in
green. Free energies of trajectories, i.e.\ $\Fen^\pif_{g}(s)$,
are denoted by the starting $s$ and goal $g$ states.  Where symbols are replaced with indices, these indices are consistent with the grid state numbering.  For example, the state in the top right corner is referred by index $6$ as clarified by \figurename \ref{fig4:infodesic-moo-12-6-b-100}C. 

The plots in \figurename \ref{fig4:infodesic-moo-12-6-b-100} relate to a near-optimal agent ($\beta=100$) navigating to a final goal in the corner ($S=\#6$). \figurename \ref{fig4:infodesic-moo-12-6-b-100}A shows the annotated heat map as values of the live distribution $\hat{p}(s;\pif)$ and the corresponding optimal policy $\pif$ as arrows.  In states on the diagonal between the goal and the opposite corner, the distribution of actions is similar to the marginalised action distribution (displayed in the yellow goal state) and are thus informationally more efficient. The policy of states not on this diagonal favours actions which are either parallel with the diagonal or move the agent towards this diagonal.  This diagram shows that the probability mass of the state distribution is maximal at state $S=\#12$ which is diagonally adjacent to the goal, with $\hat{p}(S=\#12;\pif) \simeq 0.2$. The next largest concentration of probability mass is $\hat{p}(S=\#18;\pif) = 0.08$, also on the diagonal.  This demonstrates that the policy guides the agent towards the informationally efficient states on the diagonal en route to the goal.  

\marginpar{
\vspace{.7cm} 
\small
\color{captionColour} 
\textbf{Figure \ref{fig4:infodesic-moo-12-6-b-100}. Infodesic} 
7 $\times$ 7 gridworld with the Moore neighbourhood and the
	goal in the corner state \#6 and $\beta=100$.
	\textbf{A} contains a heat map showing the live state distribution, with the
	policy distribution denoted by arrows of length proportional
	to $\pif(a|s)$ in the direction of the action.  \textbf{B} shows the proportion of sampled sequences, comprised of contiguous states, for an agent following a single policy, $\pif$, from $S=\#0$ which pass through various states en route to the final goal $S=\#6$. \textbf{C} provides a lookup grid with states labeled with their indices and sequence states highlighted in green. The
	deviation from the triangle inequality is given by the normalised free energy difference which is $-0.0005$.  We observe informationally efficient states on the diagonal, furthermore, the policy guides the agent towards these states even if it requires the agent first navigating away from the edges.  \textbf{D} shows the proportion of subgoaled sampled sequences, comprised of contiguous states, for an agent following a subgoal policy, $\pifi{1}$, from $S=\#0$ which pass through various states en route to the subgoal $S=\#12$.
}

An example for a cost $\varepsilon$-infodesic is the sequence
$\psi_{s_0\rightarrow g}= \langle \#0, \#12, \#6\rangle$ starting from
corner state $s_0=\#0$ and ending in another corner along the same
edge ($g=\#6$).     Its normalised free energy
difference $$\frac{\left(\Fen^{\pifi{1}}_{12}(0) +
    \Fen^{\pifi{2}}_{6}(12) \right
  )-\Fen^{\pif}_{6}(0)}{\Fen^\pif_{6}(0)}= -0.0005,$$ in other words,
practically respecting the equality \eqref{eq:infodesic}; this thus approaches an infodesic as defined in \ref{eq:infodesic}. We sampled actual paths, comprised of contiguous
states, of an agent following $\pif$ en route from the left corner
($S=\#0$) to the goal in the right corner ($S=\#6$), the proportion of
these sequences which pass through each state is shown in \figurename
\ref{fig4:infodesic-moo-12-6-b-100}B.  This confirms that, in
agreement with the live distribution in \figurename
\ref{fig4:infodesic-moo-12-6-b-100}A, the majority of paths visit the
state diagonally adjacent to the goal ($S = \#12$), however, as multiple $\varepsilon$-infodesics exist between state and goal pairs, it is not
guaranteed that following a single policy will hit the intermediate
state of a particular infodesic, this requires subgoaling as shown in the sampled paths of an agent following subgoal policy $\pifi{1}$ in  \figurename \ref{fig4:infodesic-moo-12-6-b-100}D.

In the simple gridworlds investigated, where the boundary serves as a
cost saving rail and the only consequence of hitting the boundary is a
delay, as information is increasingly more constrained we have
observed that corner states move closer towards each other in terms of
free energy. We therefore look at the prevalence of corner states as
interim states in cost infodesics, that is, when it is expeditious to
have the agent move first to a corner before continuing to the final
goal. It is important to note that this is a property of the present
cost regime. If instead a substantially larger penalty were applied,
then walls and corners would be avoided, especially for low values of
$\beta$ since with less information processing, the behaviour is more
more erratic. 

\vspace{.5cm} 
\vspace{-0.5cm}
\begin{adjustwidth}{-2.0in}{0in}
	\small
	\color{captionColour}
	\textbf{Figure \ref{fig5:infodesic-moo-6-18}. Corner states as interim states in infodesics.} The environment is a 7$\times$7 gridworld with a Moore neighbourhood and trade-off value $\beta=0.07$.    \textbf{A} Infodesic sequence $\langle \#38, \#6, \#18\rangle$, with normalised free energy {$\left(\Fen^{\pif}_{18}(38) - \left[\Fen^\pifi{1}_{6}(38) + \Fen^\pifi{2}_{18}(6) \right ]\right)/{\Fen^\pif_{18}(38)}= -0.187$}.  \textbf{B} \#0, \#6, \#12, \#18 with free energy $\left(\Fen^\pif_{18}(0) - \left[\Fen^\pifi{1}_{6}(0) + \Fen^\pifi{2}_{12}(6) +\Fen^\pifi{3}_{18}(12) \right ]\right)/{\Fen^\pif_{18}(0)}=-0.197$.   \textbf{C} shows a heat map of the number of times a state participates as an interim state in a 3-state $\varepsilon$-infodesic of the form $\psi_{s_0 \rightarrow s_g} = \langle s_0 = s, s_1 = s_i, s_2 = s_g \rangle$ with $\varepsilon < 0.05$.  The cell annotations show the numbering of the states.
	\vspace{.5cm}
\end{adjustwidth}

\figurename \ref{fig5:infodesic-moo-6-18}A and B show
two near-pure infodesics (with almost open-loop policies) for an agent
with restricted information processing ($\beta = 0.07$). The goal in
both cases is diagonally adjacent to the middle of the grid
($S=\#18$). For the infodesic displayed in \figurename
\ref{fig5:infodesic-moo-6-18}A, a starting state was arbitrarily
chosen on the opposite side of the grid ($s_0 = \#38$). For the second
infodesic, \figurename \ref{fig5:infodesic-moo-6-18}B, the starting
state is a nearby corner state ($s_0 = \#0$). In each case it is
informationally cheaper to have the agent first move into the corner
adjacent to the goal ($s' = 6$). \figurename
\ref{fig5:infodesic-moo-6-18}B extends the infodesic to travel
outwards along the central diagonal to the final goal,
$\psi_{s_0\to g} = \langle \#0, \#6, \#12, \#18 \rangle$.

In general the agent utilises the corners as cheaper waypoints. In \figurename \ref{fig5:infodesic-moo-6-18}C
we introduce a two-dimensional histogram in the form of a heat map which shows number of times each
state occurs as an interim state in an infodesic; we here consider all
$\varepsilon$-infodesics with $\varepsilon < 0.05$ of the form
$\psi_{s_0\to g} = \langle s_0=s, s_1=s', s_T=g \rangle$ with
non-repeating elements. The counts show that corner states, and to a lesser extent states along the edge or on the diagonal often participate as states in $\varepsilon$-infodesics. The agent is guided
by optimal policies to move to the best vantage point with respect to
information, a position where as many states as possible are
informationally closer. This is consistent with the tendency of
cognitive geometry, in our simple gridworlds with constant step penalties, to place corner and edge states more centrally for low information processing capacities (\figurename \ref{fig3:mds-moo-11-11}B and C). 

We now proceed with restricting information processing to the nearly
open-loop policy regime, $\beta = 0.01$ (\figurename \ref{fig6:infodesic-moo-0-8-1}).  Note that one characteristic of such an open-loop scenario is to be able to reach all relevant goals with a single action distribution, which is the analogue of direction in our context.

\marginpar{
\vspace{.7cm} 
\small
\color{captionColour} 
\textbf{Figure \ref{fig6:infodesic-moo-0-8-1}.  Approaching open-loop policies with $\beta=0.01$.} 
7 $\times$ 7 Moore gridworld with near-minimal information processing,
$\beta=0.01$.  The agent starts in the corner, \#0 and
aims to reach the adjacent state \#1.   \textbf{A}
Graph plot showing the $\varepsilon$-infodesic, $\psi_{0 \to 1} =
\langle \#0, \#8, \#1\rangle$, highlighted in yellow. \textbf{B} shows
the policy and Decision Information for the final goal state \#1 and
\textbf{C} shows the same gridworld policy and goal with free energies as annotations and heatmap. \textbf{D} shows the policy and free energies for the interim goal \#8. 
}

\figurename \ref{fig6:infodesic-moo-0-8-1}A shows the transition graph
of the gridworld highlighting three states of the infodesic in green.  The
agent starts in the corner ($S=\#0$) and is required to navigate to
the adjacent state $g=\#1$.   The
optimal policy to reach the adjacent final goal state is represented
by the arrows in \figurename \ref{fig6:infodesic-moo-0-8-1}B with
Decision Information as heat map and annotations.  Since the
transition function is not stochastic, the low tolerance for
information cost, $\beta =0.01$, means that the marginal action
distribution has to generalise for as many states as possible. Hence,
against naive expectation, significant probability is invested into
actions moving away from the goal.   The policy in state $S=\#0$ retains a chance of moving out of the corner
to the goal, although it is so small it is not visible on the plot,
$\pi(\shortarrow{6}|S= \#0) \approx 0.016$.  The informational cost
arising from this small difference in action distribution is clearly indicated in \figurename \ref{fig6:infodesic-moo-0-8-1}B where $\ItgDpif_{1}(0) \approx 0.15$ bits, in contrast to the significantly lower Decision Information values in all other states. 

The chance that the agent successfully leaves the corner in any given
time step is very low as it predominantly chooses one of the
prevailing actions in the policy ($\shortarrow{2}$ or
$\shortarrow{7}$), hits the edge, pays the cost of $-1$ in performance
each time and remains in the corner without  moving closer to the
goal.   This increased duration cost  combined with the increased information cost results in a significantly higher free energy, $\Fen^\pif_{1}(0) = 73.9$, as shown in \figurename \ref{fig6:infodesic-moo-0-8-1}C which displays the same policy for the same goal ($S=\#1$), but now annotated with free energy values.   

It is both more performant and informationally more efficient to
permit the agent to utilise an additional policy to facilitate leaving
the initial corner state.  To demonstrate this, we introduce an
$\varepsilon$-infodesic where the interim state is diagonally adjacent
to the initial corner ($S=\#8$) as shown in \figurename
\ref{fig6:infodesic-moo-0-8-1}D.  The agent uses one policy for the
intermediate subgoal ($s_{\text{int}}=\#8$, policy displayed in
\figurename \ref{fig6:infodesic-moo-0-8-1}D) and another policy from
there to the final goal ($g=\#1$,  displayed in \figurename
\ref{fig6:infodesic-moo-0-8-1}C). This cumulative free energy, when
using two policies $\Fen^\pifi{1}_{8}(0) + \Fen^\pifi{2}_{1}(8) =
10.92$, gives a drastic reduction of 85\% in comparison to the free
energy for using a single policy, leading to a quite
substantial violation of the positivity of the triangle relation:

\begin{equation}
  \label{eq:reduction-through-split}
  \frac{\Fen^\pif_{1}(0) - \left(\Fen^\pifi{1}_{8}(0) + \Fen^\pifi{2}_{1}(8)
    \right)}{\Fen^\pif_{1}(0)}= -0.85\;.  
\end{equation}
The discussion will deal with the ramifications of introducing such switching policies.

\section{Discussion}

We first considered the concept of geometry in the sense of a collection of geodesics which derive from a quasimetric and how they relate to the states of the given space and to each other. We then generalised this concept to the  weaker notion of infodesics which take into account the cognitive costs. The discussion that follows develops the argument that the consideration of policy switching and that of infodesics are each 
valuable in their own right.  Ignoring the cost of maintaining multiple
policies, policy switching enables agents to plan policies on subsets
of the state space instead of the entire state space at reduced
information processing costs.   Our take on infodesics is to  consider them as a collection of sequences of states, e.g.\ landmarks which emerge even in systems that do not have obvious natural choke points (doors, etc).  We posit that this collection of infodesics can be used as a sparse representation of an environment. 

In the traditional view of Euclidean geometry, one considers its
structure which binds the various aspects together and enforces
considerable constraints on its various components. In the case of
straight lines, the geodesic in the Euclidean case can be defined by
two distinct points in the space. Two such lines, if not parallel,
define intersection points, and each of these intersections then
defines an angle. Additionally, in Euclidean geometry one can ignore
the directionality of straight lines due to their directional
symmetry. When moving to Riemannian geometry, a few of these
constraints (such as the sum of angles in a triangle) are relaxed or
dropped entirely. In the case of a Finsler geometry, the underlying
metric is further relaxed into a quasimetric and the directionality of
the geodesics becomes important owing to this asymmetry.

While these spaces can be derived by metrics or quasimetrics, most of
these spaces are not considered primarily in terms of distances, but
in terms of a collection of the geodesics and how different geodesics
relate to each other. However, here we adopted the perspective that,
whatever structure one wishes to consider, it must derive from the
costs of solving a task. In the case of deterministic value geodesics,
this essentially reduces to deriving the latter from a quasimetric.

Such considerations become especially important when moving to
resource-rational approaches where cognitive costs matter. Being able
to maintain a behaviour along a trajectory makes the task solution
along a sequence of states cognitively cheaper, even if the pure
distance cost may differ.  When travelling to a goal, one has two components of
cognitive operation. For the first, the policy, against which the given
state needs to be monitored, secondly, some identificator trigger or cognitive
feature that signals that the goal has been reached and the agent can
stop.

Now, a point in favour of ``desics'' is that there may be a suitable
sensor or cognitive feature which will keep one on a ``desic'' (e.g.\
keeping a compass direction, a fixed angle towards the sun or policy).
In contrast, the ``hitting'' of a goal state can be outsourced to
another decision-making entity, e.g.\ a timer that estimates how much
time remains before arriving at the goal, or some sensor triggered by the
goal state, or, as in our case, by making the goal state (or subgoals,
in the case of trajectory segments) explicitly absorbing. This means
that following the desic itself is a low-information process as one is
not required to keep track precisely of how far down a route one has
moved. Stopping once a target is hit is considered a separate process,
either handled by a separate cognitive or sensoric unit or by a
trapping state of the environment itself. In other words, we
separate this decision-making into a processual component and an
identificatory component.

Before being able to capture that, we had first to adapt definitions to
discrete spaces.  So we proceeded to define geodesics by state
sequences whose subsegments are distance-wise consistent with the
overall sequence, i.e.\ which turn the triangle inequality into an
equality. This works seamlessly for deterministic value geodesics. However, with the introduction of cognitive costs, when optimal policies can become probabilistic, the infodesic criterion based on the triangle inequality becoming an equality does not transfer smoothly. Instead, one obtains strongly impoverished infodesics, unless one relaxes the constraint. This, while violating the desirable properties associated with metrics or quasimetrics, has some instructive implications which we discuss below.

Any structured strategy-finding reflects the intrinsic structure of
the state space and its coherence rather than just carving out a
single isolated trajectory from a specific starting state to a
specific single goal. Understanding the structure of a task space
thus inherently adopts a multi-goal perspective. To discuss its
validity in predicting the decision-making behaviour of a
resource-rational agent, consider, for instance, a person navigating
to a destination while distracted or multi-tasking, perhaps they are
talking on their phone while walking. In line with recent work on
simplified mental abstractions \citep{Ho2022:mental-representations}
our expectation is that people will choose a less demanding route that
will enable them to pay less attention to their exact location, and
therefore plan a policy which can be more generally applied. Akin to
compressing the state space, instead of storing a separate action
distribution for each state, we can cluster the states and select a
more universal policy which reduces
the cost of selecting the appropriate actions in the different states.
We can then ask which other goals can be achieved either equally
optimally or with little extra effort with this particular action
selection model, and extending this to the idea of agents switching
goals on the fly while expending little additional computation.

When one generalises from the problem of reaching a particular goal in
a cost-only MDP to a general set of goals, one makes the transition
from finding optimal trajectories with a specific end state to a whole
arrangement of optimal sequences across the space. The idea remains
the same. ``desics'' are in this perspective those sequences of states
that are optimally reachable from the starting point. Geodesics are
the most convenient of such collections as they have a metric or
quasimetric from which they can be derived. However, in this
perspective, the ``desic'' (geo- or info-desic) view does no longer
require them to be actually derived from a strict metric or
quasimetric, although, for better interpretability we still strive to
derive it from a concept that is as close to the quasimetric as
possible.

\subsection{Generalising Geodesics}\label{generalising-geodesics}

When thinking in terms
of geodesics one can, instead of considering it an optimal route
between two states in the space, select these by choosing a starting
state $s$ and a generalised ``direction''. We reiterate that in our framework we
generalise the choice of a fixed compass direction to the choice of an
open-loop policy. Following that policy fixedly will keep us moving in
a particular direction towards the goal. This is best illustrated by
way of an example: when humans navigate they often abstract the route;
for example, assume that the shortest path runs along a fixed compass
direction, say, North. If they later find they need to also visit a
waypoint which happens to be between the starting point and the goal,
they  profit from the fact that their selected policy is also optimal
for any waypoints in between. This means that all states on that
geodesic, and not just $g$, are reached optimally from $s$ under this
policy. Identifying geodesics determines an optimal trajectory of goal
states along the way and thus solves a whole class of problems at
once, which is a critical motivation for the geometric perspective.

We point out a dichotomy between considering ``shortest paths''-type
and ``constant directions''-type criteria. The shortest path route is
attained by the optimisation of the value term in the free energy,
while the optimisation of the Decision Information term corresponds to
minimising the state-specific deviation of the action distribution
from the overall marginal. The latter can be seen as an attempt to
make the policy approach an open-loop policy as closely as possible.
In this interpretation, cognitively cheap routes correspond to the
direction-preserving trajectories. This means that one can interpret a
free energy optimisation at given $\beta$ as selecting a trade-off between
shortest-route and direction-preserving ``desics''.


We can obtain geodesics based on shortest routes from the optimal
trajectories between a state $s$ and a goal $g$ whenever they are
derived from a value function and executed as deterministic
trajectories. On the one hand, the Bellman property guarantees that
for any intermediate state $s'$ on such a trajectory the
\emph{remaining} part of the trajectory from $s'$ to $g$ must be
optimal; for a proper interpretation as a geodesic, we also require
the counterpart: the \emph{preceding} portion leading from $s$ to $s'$
should be optimal.

We propose that this perspective throws a different light on the
organisation of tasks, namely, instead of considering the task
solution simply as a ``recipe'' that is carried out starting in a
state to reach a goal, with the Bellman equation merely providing a
computational advantage, it rather defines a similarity and an overlap
of the ``basin of attraction'' of nearby goals. If one considers its
dual, namely the ``zone of repulsion'' around the starting state, this
leads naturally to the concept through which we define the geodesics:
namely that the \textit{initial} segment ($s$ to $s'$) of a
geodesic must also be optimal. This is the motivation for our
definition of the generalised geodesic as one where the (generalised)
triangle inequality \eqname \eqref{eq:geodesic} must hold as an
equality for a sequence of the elements of the geodesic.

In analogy to the basin of attraction which creates concepts of
similarity for goals via the Bellman property, the zone of repulsion
thus creates such a similarity for starting states via the requirement
of optimality for the initial segment. In other words, one obtains a
natural structure of task similarities and a set of relations with
respect to each other, rather than treating the task of reaching a given goal as a completely different undertaking for each distinct goal.  Additionally, the concept of  geodesics creates a class of goals which are solved by maintaining the same policy. This means that, even if the goals are not close to each other, they belong in the same class of problems being solved by the same policy. 

This criterion renders the totality of the elements of the sequence
following this policy optimally reachable from the starting point.
With added information processing constraints in the form of Decision
Information, this generalises the existing concept of pure value
geodesics to \emph{infodesics}.  By using this term
we imply the totality of the infodesics, their relation to the various
states in the state space, and with respect to each other. This creates characteristic
distortions in the geometry of the problem under consideration and
defines what we call our \emph{cognitive geometry}. 

A cognitive geometry introduces some novel effects. With reduced information processing capacity, optimal policies typically become stochastic, which can render infodesics trivial, i.e.\ only $s$ and $g$ themselves fulfil the triangle equality criterion. We therefore included relaxed versions of the equality to still be able to consider near-infodesics. However, further drastic deviations are possible, such as in \eqname \eqref{eq:reduction-through-split}, where splitting a trajectory into two parts governed by different policies can significantly \emph{reduce} the total cost. This corresponds to switching policies at the intermediate state $s'$. The latter means that one uses two different marginal action distributions throughout the run and thus enables the execution of a more complex overall policy than with a single trajectory. Notably, and as a new phenomenon, this reduction in information only appears because the original triangle inequality does not include any complexity cost involved in a policy switchover in Decision Information, despite the fact that this requires actual cognitive effort. We thus expect that a complete characterisation of cognitive geometry will need an expanded form of the triangle inequality that will take such policy switching costs into account in a systematic way.

\subsection{Free energy as a Quasimetric?}

Ideally one would have liked the infodesics to derive from a quasimetric. When we consider free energy against the requirements for a quasimetric space, we find the following:

\begin{enumerate}[label={D\arabic*}:]
	\item Nonnegativity:  we have already shown that $\Fen^\pif_g(s)\geq 0$ as $V^\pif(s) \leq 0$ and $\ItgD^\pif(s) \geq 0$, where $\pif$ is optimised in terms of free energy for goal state $g$.
	\item Principle of Indiscernibles: we know that $\forall g \in \G ~
          \Fen^{\pi}_g(g)=0$ for all $\pi$ as goal states are absorbing and for $ s \in \S \setminus g, ~ \Fen^\pi_g(s) \neq 0$ for all $\pi$.
	\item Asymmetry: In general, the optimal policy navigating
          from $s \stackrel{\pifi{1}}{\longrightarrow} g$ differs from the policy navigating
          from $g \stackrel{\pifi{2}}{\longrightarrow} s$.  Therefore, $\ItgD^\pifi{1}(s) \neq
          \ItgD^{\pifi{2}}(g)$ and  hence  $\Fen^\pifi{1}_{g}(s) \neq
          \Fen^{\pifi{2}}_{s}(g)$ in general, making free energy at best a quasimetric.
	\item Triangle inequality: we ask whether $\Fen^\pi_g(s) \stackrel{?}{\leq}
          \Fen^{\pifi{1}}_{s'}(s) + \Fen^{\pifi{2}}_g(s') ~ \forall s \in \S$, where
          each segment utilises a different optimal policy $\pifi{1}$
          and $\pifi{2}$.  
\end{enumerate}
For the free energy to form a quasimetric, the crucial axiom to establish is that the triangle inequality holds. However, we have already demonstrated with a counterexample that the partitioned cumulative free energy can be lower than the free energy without subgoaling \eqref{eq:reduction-through-split}. Hence, in general, the triangle inequality does not hold and the free energy is not a quasimetric.

\subsection{Task Decomposition}

\textit{Rational analysis} involves understanding human cognition as a rational adaptation to environmental structure and it is extended by \textit{resource-rational analysis} which aims to explain why people may choose to adopt particular goals or heuristics in view of their cognitive resources \citep{Lieder2020:resource-rational-cognition}.   People's cognitive strategies are jointly shaped by the environment and computational constraints when planning strategies to complete a task \citep{Callaway2018:resource-rational-human-planning}.  \citet{Maisto2015} proposes that, during planning, intermediate goals are selected to minimise the computational Kolmogorov complexity.  Resource-rational task decomposition, which reduces the computational overhead of planning at a primitive action level, is consistent with elements of previous studies on human planning  \citep{Correa2020:resource-rational-planning}.   \citeauthor{Ho2022:mental-representations} conducted trials of candidates navigating a simple gridworld through a variety of obstacles to reach a goal which demonstrates that people plan by constructing a simplified mental representation of the task environment \citep{Ho2022:mental-representations}.  Here we propose a complementary interpretation, where making use of segmentation by incorporating interim goals one reduces informational costs of the individual infodesic routes. 

In our context, consider the classical mountaineering techniques known as ``aiming off'' and ``handrailing'' \citep{Langmuir2013}, which are often combined in poor visibility.  Aiming off entails subgoaling in the form of deliberately aiming away from the intended goal instead of trying to walk on an exact bearing,  for example, aiming due East of the goal.  Handrailing is when the mountain leader walks along  a feature of the landscape, i.e. a cliff, fence, etc., this provides robust and cognitively cheap navigation.   For example, an interim goal is selected which is known to be due East along a fence from the actual intended destination.  The mountain leader aims off to the interim goal which is known to be due East of the final goal, arriving at this interim goal, one then knows the final goal is due West along the fence.  The mountain leader then handrails the fence to arrive at the final destination.   This task decomposition enables a person to benefit from a reduction in the information processing required due to the structure of the environment, i.e.\ an epistemic strategy:  physical actions ``that [make] mental computation easier, faster or more reliable'' \citep{Kirsch1994}.  

When segmenting an infodesic, one decomposes the original problem into
subproblems with given subgoals. The constituent free energies for
these subgoals utilise different policies. In the deterministic case
with greedy value optimisation ($\beta \to \infty$), no state will be
revisited twice throughout a trajectory. This means that we can always
construct a superpolicy for the overall goal-seeking behaviour by
integrating the respective subpolicies and policy-switching is not
required. Even if we consider lower $\beta$ values, whenever there is
no ambiguity about which subpolicy applies at which state, i.e.\ no
overlapping of the support for the different policies, the validity of
the triangle inequality for free energy can be upheld, as shown in the
theorem presented in \ref{S-triangle-inequality-disjoint-policies}.
However, when we reduce $\beta$, policies typically become
increasingly stochastic to reduce the divergence from the action
marginal. At some point, the stochastic trajectories belonging to
different subgoals will begin to overlap en route: the agent may visit the same state during different segments of the infodesic, and therefore under different policies.   At the given state it is thus now no longer uniquely resolvable without further information which of the subsegment policies is currently active.  

As we have shown, although without taking any switching costs into
account, this enables the agent to benefit from informationally more
efficient policies,  having an advantage over single policies by
better making use of the structure of the environment (similar to
temporal abstractions in \citep{Sutton1999} and subgoaling in
\citep{Dijk2011a}).  For instance, as can be seen in the example shown
in \figurename \ref{fig5:infodesic-moo-6-18}B, by moving to the corner
opposite the goal, the agent aligns all future actions with the
marginalised action distribution thereby reducing the cost of
information processing even while incorporating the sequence $\langle
\#0, \#6, \#12, \#18 \rangle$.  To ascertain which is cognitively
cheaper will require a more thorough understanding of the switching
costs.  It is possible that there are cases when it may be cheaper to
use multiple policies even when taking the switching costs into account.  

\subsection{Grid Cells and Cognitive Maps}

It is becoming increasingly central to study the geometry of perception and decision-making \citep{Chung2021, Kriegeskorte2021}, thus, one reason to strive towards a formalisation of cognitive geometries is the long-standing idea that humans and animals form cognitive maps of their environment \citep{tolman1948cognitive}. A large body of neurophysiological work has demonstrated that the hippocampus is sensitive to the geometry of the environment, suggesting that it may support the formation of cognitive maps \citep{o1978hippocampus}. Two cell types are of particular interest in this: place cells which fire at unique locations in the environment \citep{o1971hippocampus} and grid cells which fire at multiple locations forming a regular, triangular lattice \citep{Hafting2005}. The former is suggested to enable topological navigation while the latter may provide a spatial metric supporting path integration and vector-based navigation \citep{edvardsen2020navigating}.

While there is strong evidence to suggest that the hippocampus is involved in spatial navigation, growing evidence suggests that the representation of location in itself may not be the sole purpose and that the hippocampal formation may instead also support other important functions, such as reward-guided learning \citep[e.g.][]{boccara2019science,butler2019science}. Since success in reinforcement learning does not rely on Euclidean distance, but the distance along paths (e.g.\ around obstacles), efficient reinforcement learning should thus rely on a geodesic rather than a plain Euclidean metric \citep{gustafson2011grid}. In a series of simulations, \citep{ gustafson2011grid} provide theoretical support for the benefit and empirical evidence of this spatial coding and hypothesise that the firing of place and grid cells may be modulated accordingly. 

In line with this work, we here additionally speculate that it is not only a purely geodesic metric that is of importance to biological agents, but that additional cognitive costs, which we model by information-theoretical quantities, should also affect computations and hence be supported by the underlying metric. Since the hippocampus is also involved in the representation of information like value \citep{knudsen2021hippocampus} and goals \citep{boccara2019science,butler2019science, CrivelliDecker2021biorxiv}, or even conceptual spaces  \citep{constantinescu2016organizing, theves2020hippocampus}, it appears that activity of cells found in the hippocampal formation may reflect geometrical properties more generally. In this light, it appears not without reason to hypothesise that this brain area may also be a good candidate to probe for its sensitivity to the geometry that arises from a trade-off between value and information as demonstrated in this work.

\subsection{Future Work}

For a complete model of the concatenation of sub-infodesics, it will
be necessary to model the detection of a subgoal being achieved,
together with a memory state of the agent that switches to keep track
of the currently active subgoal. We expect an informational overhead involved
with the policy switching costs $C_\text{switching}$ for this
infodesic composition needs to be incorporated in the overall costs.
We conjecture  $C_\text{switching}$ to be most aptly modeled as a cost that captures the
information indicating which subgoal is currently active, or, more
precisely, the cumulated cost of checking the  currently active
policy, for each decision taken, since we have a memoryless agent.  We
posit that, while the total decision complexity is reduced by
switching  between multiple policies, when a suitably defined cost
$C_\text{switching}$ of maintaining these multiple policies is
included, we will obtain a ``relaxed'' triangle inequality of the form: $\Fen^\pif_g(s) <= \Fen^\pifi{1}_{s'}(s) + \Fen^\pifi{2}_g(s') + C_\text{switching}$, which takes this switching cost into account and would be the necessary condition for using the free energy as a relaxed quasimetric.   This switching cost might be characterisable systematically from the properties of the task space. If verified,  it would be an interesting target for investigation to study  how switching costs  could contribute to identify particularly cognitively convenient states in the task space.

Information-to-go is only one very specific measure and it might be insightful to study other resource-rational measures in a geometric context instead. This will possibly require other adaptations of the triangle inequality. These could provide ways of comparing possible preferred behaviour segmentations under these measures with experimental observations.

Our scientific objective is to use this framework as a model for the way in which bounded rational agents, human and robots, map their knowledge of interrelations between states, in the context of navigation this would be how to travel efficiently between known locations using previously planned optimal routes. An additional experiment that would allow one to confirm the cognitive plausibility of this framework is inspired by the trials conducted by \citeauthor{Ho2022:mental-representations} \citep{Ho2022:mental-representations}. The experiment involves a navigation task in a virtual maze solved by human subjects under information processing constraints, which are induced by requiring that candidates recall their routes at the end of the task.  Candidates will be surveyed regarding their estimation of the distances between pairs of states in the environment.   Our hypothesis is that once the experiment has been conducted, the distances estimated by the candidates will correspond to the distances predicted by the proposed cognitive geometry.  

When humans plan a route, they often do so with previously experienced
and memorised routes in mind and utilising these as points of
reference to explore unknown territory. Then, upon implementing the
plan, they assess the route and depending on their evaluation they
either store the information to be used again or discard the route. In
other words, we have the implicit assumption that the ``desics'',
rather than some random routes, serve as a natural skeleton from which
to look for ways to explore the unknown parts of the space.

\section{Conclusions}

We considered distances induced by cost-only MDPs which were additionally endowed with an informational cost reflecting the complexity of decision-making. Geodesics generalise the intuition about geometry determined by directions and distances, representing optimal transitions between states. We proposed that the addition of informational criteria would characterise a \textit{cognitive geometry} which additionally captures the difficulty of pursuing a particular trajectory. 

We found that free energy retains some of the structure of the spatial geometry via the value function while incorporating the cost of information processing. However, emphasising the most informationally efficient policy amongst otherwise equivalent policies will favour trajectories which pass through informationally efficient states en route to the goal. Such trajectories therefore often experience a ``detour'' in terms of pure distance through more easily maneouvrable hubs.  In specific examples we found considerable distortions which place boundary states more centrally in the space, with the boundaries acting as guides. Additionally, when  considering infodesics, i.e.\ sets of intermediate states which are optimally reachable from the starting state, intermediate goals can be achieved en route  to the final goal, thus defining classes of problems that are solved as a side effect of solving the main one.

Analogously to geodesics, we characterised the infodesic property by the triangle inequality becoming an equality. Since this inequality is not always perfectly respected in our framework, we had to relax the conditions.  Furthermore, due to the informational nature of the free energy
distance, splitting a trajectory devolves the possible cost of
switching the policies of the two segments into the split itself.
This can cause the violation of the triangle inequality to be quite
significant. In the future, we aim to explicitly incorporate the
splitting cost into a generalised, but tighter triangle inequality
for a more stringent description of the infodesic. Note that one can
separate the contributions of the trajectory cost and the splitting
cost and thus to understand (and control) how much each of these
contributes to trace how much of the complexity is found in the
trajectory and how much in the splitting.
In the future, the splitting cost will be explicitly incorporated into a generalised, but tighter triangle inequality for a more stringent description of the infodesic.

Establishing such a cost, we propose, will allow us to impose, with its quasi-distance structure and additionally the ``directional'' structure induced by the policy choice, a quasi-geometrical signature on the state space. We suggest that this offers the basis of a genuinely geometrical notion of task spaces that takes into account cognitive processing: a cognitive geometry. This we understand to be a structure with optimal trajectories determined by either two states or by one state and a ``direction'' (i.e.\ policy) that is informed not only by the pure spatial geometry, but also by the cognitive costs that an agent needs to process when moving from task to task and how it has to informationally organise policies to achieve nearby or related tasks. Critically, such a concept would suggest that straightforward use of Euclidean, or geodesic-based geometry may not be the appropriate language to treat even purely navigational decision problems. For that reason, it would be interesting to investigate this distance further.

\section*{Acknowledgements}
We would like to thank Dr Giovanni Pezzulo for pointing out relevant literature and for his insights when discussing possible experiments to enable verification of the plausibility of our framework.

In addition, we would like to thank the reviewers for investing their
time and effort in providing very detailed comments which lead to a considerable improvement in the clarity and quality of the manuscript.
 
\setcounter{figure}{0}
\setcounter{equation}{0}
\setcounter{section}{0}
\renewcommand{\thefigure}{S\arabic{figure}}
\renewcommand{\theequation}{\roman{equation}}
\renewcommand{\thesection}{S\arabic{section}}   

\vspace{2em}
{\section*{Supporting Material}}

\section{Code Repository}\label{S-code-repository}
All code and data (including Jupyter notebooks to reproduce the simulations) are available from the University of Hertfordshire's Adaptive Systems Research Group repository via \url{https://gitlab.com/uh-adapsys/cognitive-geometry/} and have been archived within the Zenodo repository: \url{https://zenodo.org/record/7273868}.

\section{Probabilistic State and Action Pairs}\label{S-state-action-pairs}
A policy $\pi(a|s)$ is a mapping between a state and action, and the conditional probability of taking that action in that state, namely $\pi: A \times S \rightarrow [0,1]$.  So, the policy in general defines a stochastic choice of actions. We then consider the distribution of all possible future trajectories parameterised by $\pi$  an agent can traverse starting in $s_t \in \mathcal{S}$ (denoted by $\Xi_{s_t}^ \pi$), following the policy $\pi$ from time $t$ until the trajectory terminates at a goal state $s_T = g$.  The realisation of a particular trajectory within this distribution is denoted by $\xi_{s_t}^\pi \in \Xi_{s_t}^\pi$.  Throughout the paper we are dealing with optimal policies and assume that the agent ultimately reaches the goal which is absorbing. Given a fully connected world with no recurrent states, the time of termination $T$ is a random variable which varies according to the length of the trajectory. Under the Markov assumption one has the probability:

\begin{align}
p(\xi_{s_t}^\pi) &= p(a_t, s_{t+1}, a_{t+1},\dots, s_{T} |s_t) \label{eq:trajectory-xi}\\
&= \prod^{T-1}_{i=t} \pi (a_i|s_i)p(s_{i+1}|s_i, a_i)\\
&=   \prod^{T-1}_{i=t} \pi (a_i|s_i) \cdot \prod^{T-1}_{i=t}p(s_{i+1}|s_i, a_i) \label{S-eq:state-action-probability-sequence}.
\end{align}

\section{Value Function}\label{S-value-function}

For a given policy, the return $R^\pi(s_t)$ for a sequence of actions, contingent on the policy $\pi$, from time $t$ to the final time step $T$ is the sum of the rewards for each time step \citep{Sutton2018}. As outlined in the main text, in this work we use cost-only MDPs with the reward function as follows: $r(g,\cdot,\cdot) = 0$ for goal states and $r(s',s,a) = -1, \quad\forall s, s' \in \{\S \setminus g\}$ for all transient states ($s_{t+1}$ is denoted as $s'$ and $s_t$ is denoted as $s$ where unambiguous, with random variables $S, S'\in \S$).  We assume policies which result in acyclic trajectories which thus terminate, which enables us to use an undiscounted return.
	\begin{equation}
		R^\pi(s_t) \defeq r(s_{t+1}, s_t, a_t) + r(s_{t+2}, s_{t+1}, a_{t+1}) + \dots + r(s_T, s_{T-1}, a_{T-1}).
	\end{equation}
The value function $V^{\pi}(s)$ represents the expected return of an agent when starting in a given state $s$ and following policy $\pi$ to the end of the trajectory:
	\begin{equation}
			V^{\pi}(s)  \defeq \E _{\pi(a|s)p(s'|s,a)}[R^\pi(s)].\label{eq:value_function}
	\end{equation}

As the return is undiscounted, the negative of the value function quantifies the number of actions taken, which we can interpret as a measure of performance of the policy $\pi$.  The Bellman equation \citep[see (3.14) section 3.5]{Sutton2018} is a recursion relation which expresses the relationship between the value of a state and the values of its successor states, see \eqname \eqref{S-eq:value-bellman}.  As an expectation over all future trajectories, the Bellman equation averages over all possibilities weighted by the probality of each possibility occurring.

\begin{align}
	V^{\pi}(s)  &=  \sum_{a}\pi(a|s)\sum_{s'}p(s'|s,a)\left[ r(s', s, a) +
    V^{\pi}(s')\right].\label{S-eq:value-bellman}
\end{align}

The Principle of Optimality due to Bellman \citep[see section 1.3]{Bertsekas2017:dyn-programming-vol-1} states that “An optimal policy has the property that whatever the initial state ($s_t$) and initial decision ($a_{t-1}$) are, the remaining decisions must constitute an optimal policy with regard to the state resulting from the first decision.” This forms the basis of dynamic programming as the problem is divided into tail subproblems which are solved in reverse time order.

Value functions define a partial ordering over policies. A policy $\pi'$ is said to be better than or equal to another policy $\pi$ if the value function indicates a better performance, i.e. $V^{\pi}(s) \leq V^{\pi'}(s)$ for all $s \in \mathcal{S}$.  There is always at least one value optimal policy $\piv$ with optimal value $V^*$ which is better or equal to all other policies.  In the main paper we talk about the policy $\pif$ which minimises free energy.  This results in the agent acting optimally, but under a given information constraint specified by a trade-off parameter $\beta$.

The value of selecting an action $a$ in a particular state $s$ and following the policy $\pi$ thereafter is given by the \textit{state-action value function} $Q^\pi(s,a)$, as defined below \citep[see (3.6) section 4.2]{Sutton2018}. 

\begin{align}
	Q^{\pi}(s,a) &\defeq \E_{p(s', r|s,a)}\left[r(s',s,a)+V^\pi(s')\right]\label{S-eq:state-action-value-function}\\
	&= \sum_{s',r}p(s',r|s,a)\left[r(s',s,a)+V^\pi(s')\right].
\end{align}

\section{Live State Distribution}\label{S-live-state-distribution}

Given the random variables $S, A \sim p(s,a)$ we can calculate the action distribution by marginalising the joint probability over all states, $p(a) = \sum_{s\in\mathcal{S}} p(a|s)p(s)$.  The conditional probability $p(a|s)$ is given by the policy $\pi$.  The state distribution $p(s)$ is often assumed to be unifomly distributed, i.e. $p(s) = \frac{1}{|\S|}$.  In this work we are estimating the cost of implementing a policy and therefore more concerned with the states actually visited by the agent while following the policy.  For this reason we employ a \textit{live state distribution} $p(s; \pi)$ which is a modified stationary state distribution for an absorbing Markov chain. We use the semicolon to indicate that the state distribution is parameterised by the policy $\pi$. 

The one-step transition probabilities are calculated by combining the policy with the state-action probabilities according to $p(s_j| s_i) = \sum_{a \in A} \pi(a|s_i) P^{a}_{s_i s_j}$ for all $s_i, s_j \in \S$, thus we have a Markov chain from the MDP. Let $\mathbb{P}$ be the transition matrix of the Markov chain, where the $ij$th entry $p_{ij}^{(n)}$ of the matrix $\mathbb{P}^{n}$ is the probability that starting in state $s_i$, the system will transition to state $s_j$ after $n$ steps.  Let $\mathbf{u}$ be the probability vector with entries $u_i$, which represent the starting distribution of states $s_i$, then after $n$ transitions, the state distribution is given by $\mathbf{u}\mathbb{P}^n$.  In our case we start with a uniform state distribution, $\mathbf{u} =
\frac{1}{|\S|-1}, i \in \{1,\dots,\vert\S\vert-1\}$ for all $s \in \{\S \setminus g\}$. 

The live distribution only takes into account processes which have not
yet reached the absorbing states, i.e.\  the probability mass in the
terminal state is ignored \citep[see section 11.2]{Grinstead2006}. It
only considers transient states, as such, recurrent and absorbing
states are given zero probability mass.   Following the method
described by \citeauthor{Grinstead2006}, we define $\mathbb{Q}$ as the
submatrix of $\mathbb{P}$ which contains only the transition
probabilities of transient states, in our case we are considering a single absorbing goal state.  Thus the entries of $\mathbb{Q}^n$ give the probabilities for being in the transient states after $n$ steps for each possible transient starting state.

For an absorbing Markov chain the matrix $I - \mathbb{Q}$ has an inverse $\mathbb{N}$, whose entries $n_{ij}$ give the
expected number of times a process starting in state $s_i$ visits
state $s_j$ before absorption. $\mathbb{N}$ is also referred to as the \textit{Fundamental Matrix} of
$\mathbb{P}$. The time until absorption starting from each state is
then given by the vector $\mathbf{l} = \mathbb{N}\cdot\mathbf{c}$ with entries $l_i$ being the ``live'' time for a trajectory starting in state $i$.   $\mathbf{c}$ is a column
vector with all entries equal to 1. Thus, $l_i$ is the sum of entries in the $i$th row of $\mathbb{N}$. Then, the live distribution $p(s_j;\pi) = \sum_i \frac{n_{ij}}{l_i}u_i$. This live
distribution is indicative of how many times the agent visits each
transient state and therefore summarises the trajectory statistics
taken by the agent by virtue of following the policy $\pi$. In
contrast, the classical (unmodified) stationary state distribution will result in  the probability mass 
being concentrated in recurrent and absorbing states as $t \to \infty$.

\section{Information Measures}\label{S-information-measures}

Shannon describes information as a measure of uncertainty \citep{Shannon1948}. If the outcome of a random variable X is known, its uncertainty is nil and observing the event gives us no new information.  As such, the uncertainty in $X$ can be measured using Shannon's entropy \citep{Cover2006}. Let $X$ be a discrete random variable with alphabet $\mathcal{X}$, $X \in \mathcal{X}$, and $X \thicksim p(x)$. The entropy $H(X)$ is defined by
	\begin{equation}
		H(X) \defeq -\sum_{x\in \mathcal{X}} p(x)\log p(x). \label{eq:entropy}
	\end{equation}

The probability mass function $\Pr\{X=x\}$ is denoted in brief as $p(x)$ by abuse of notation, whenever its meaning is unambiguous.  Here, the logarithm is taken with respect to base 2 and  information is measured in bits. We also apply the usual convention of setting $0 \log 0 = 0$ as by continuity of $x \log x \rightarrow 0$ for $x \rightarrow 0$ from above.

Relative entropy, or the Kullback-Leibler divergence $D_{KL}$ \eqname \eqref{eq:D_KL}, is a measure of how much additional information is necessary to code according to an assumed distribution $q(X)$ as compared to when one would assume the correct distribution $p(X)$. Although it is often interpreted as a distance measure between two probability distributions, it is not a true metric between the distributions because it is not symmetric and does not satisfy the triangle inequality \citep{Cover2006}. 
\begin{equation} 
	D_{KL}[p(X)||q(X)] = \sum_{x\in \mathcal{X}}p(x)\log \frac{p(x)}{q(x)}\label{eq:D_KL}. 
\end{equation} 
Extending the continuity argument above, we adopt the convention that $0\log\frac{0}{r} = 0$ and $s \log \frac{s}{0} = \infty$ for all $r \geq 0$ and $s >0$.

Mutual Information is a measure of the mutual dependence between two variables, as such it is the KL divergence between their joint distribution and the product of their marginal distributions, see equation \ref{eq:mutual_information_KL}.  Like all KL divergences, mutual information is always nonnegative, in addition it is zero when the joint distribution coincides with the marginals (i.e., where the two variables are independent).

\begin{align}
	I(X; Y) = I(Y; X) & = D_{KL}\left[p(X,Y)||p(X)p(Y)\right] \label{eq:mutual_information_KL}\\
	&= \sum_{x\in \mathcal{X}}\sum_{y\in \mathcal{Y}}p(x,y)\log \frac{p(x,y)}{p(x)p(y)}\label{eq:mutual_information}
\end{align}

If we consider a state-action information channel, then the mutual information $I(S;A)$, as shown in \eqref{eq:E-DKL-policy-actions}, can be shown to be the expected KL divergence of the conditional probability of an action given a state $p(a|s)$, which is synonymous with the policy $\pi(a|s)$, and the action distribution $p(a)$.  This is the amount of information that an agent needs to acquire from the environment to select an action.

\begin{align}
    I(S;A) & = \sum_{s\in \mathcal{S}}\sum_{a\in \mathcal{A}} p(s,a) \log \frac{p(s,a)}{p(s)p(a)} \label{eq:DKL-states-actions}\\
    & = \sum_{s\in \mathcal{S}}p(s)\sum_{a\in \mathcal{A}} p(a|s) \log \frac{p(a|s)}{p(a)} && p(S,A) = p(A|S)p(S)\\
    & = \sum_{s\in S}p(S) D_{KL}(p(A|s)||p(A)) \label{eq:E-DKL-policy-actions}
\end{align}

\subsection{Relevant Information} 

\textit{Relevant Information} \citep{Polani2001} is defined as the minimum mutual information $I(S;A)$ that an agent needs to process when selecting an action within a state to obtain an average utility equal to a given threshold ($\bar{U}$).  The constrained optimisation is solved via the Lagrangian equation \eqref{eq:relevant-information-L}, where the last term corresponds to the normalisation constraints.  The Lagrangian multiplier $\beta$ specifies the trade-off between utility and mutual information.  
\begin{equation}
	\min_{\pi(a|s)} I(S;A) \text{ s.t. } \E_{p(s, a)}[Q^\pi(s, a)] = \bar{U}\label{eq:relevant-information-L}
\end{equation}

\begin{equation}
	\mathcal{L}^\pi(\beta)=I(S;A)+\beta\left(\bar U -\sum_{s, a}\pi(a|s)p(s) Q^{\pi}(s,a)\right) + \sum_s\lambda_s\left(1-\sum_{a}\pi(a|s)\right).\label{eq:relevant-information-L-con}
\end{equation}

The algorithm to calculate Relevant Information \citep{Dijk2013} optimises this Lagrangian to solve for the particular policy which minimises the mutual information $I(S;A)$ while achieving an average action value of $\bar{U}$.  Relevant Information is the average KL divergence between an average action that minimises the Lagrangian and a prior action distribution, see \eqref{eq:E-DKL-policy-actions}. In essence, it is a single decision version of Information-to-go, which we present in the next section.

\subsection{Information-to-go}\label{S-information-to-go}

\textit{Information-to-go} \eqref{eq:information-to-go} extends Relevant Information to a sequence of decisions.  It quantifies the information processed by the whole agent-environment system over the course of the state-action sequences.  It is the KL divergence between the joint probability of future trajectories conditioned on the current state and action, and a fixed or estimated prior. It is the information required to enact the trajectory by the whole agent-environment system as compared to the prior and can be decomposed into separate components for decision complexity and for the response of the environment \citep{Tishby2011:information-to-go}. As a Kullback-Leibler divergence, Information-to-go is non-negative \citep{Cover2006}:  

\begin{equation}
	\Im^{\pi}(s_t,a_t) \defeq \E_{p(s_{t+1},a_{t+1},s_{t+2},a_{t+2},\dots|s_t,a_t)}  \left[\log \frac{p(s_{t+1},a_{t+1},s_{t+2},a_{t+2},\dots|s_t,a_t)}{\hat{p}(s_{t+1},a_{t+1},s_{t+2},a_{t+2},\dots)}\right]. \label{eq:information-to-go}
	\end{equation}


As we are investigating geometry, we wish to quantify the Information-to-go of a path from a given state to a target state, without considering an initial action and therefore, we do not condition on the current state and action pair, but rather on the current state only, denoted $\bar{\Itg}^\pi(s), $see \eqref{eq:information-to-go-state}.  Where the decoration $\hat{p}(\cdot)$ denotes a prior probability distribution.
\begin{equation}
	\bar{\Im}^{\pi}(s_t) \defeq \E_{p(a_t, s_{t+1},a_{t+1},s_{t+2},a_{t+2},\dots|s_t)}  \left[\log \frac{p(a_t,s_{t+1},a_{t+1},s_{t+2},a_{t+2},\dots|s_t,a_t)}{\hat{p}(a_t,s_{t+1},a_{t+1},s_{t+2},a_{t+2},\dots)}\right]. \label{eq:information-to-go-state}
	\end{equation}

Using the linearity of expectation, the rules of logarithms and the factorisation of the joint probabilities of independent events, we can separate a term in \eqref{eq:information-to-go-sepd} resulting in a Bellman-like recursive equation for state only Information-to-go \eqref{eq:information-to-go-bellman}.  The expectation is again over the future trajectories originating from state $s_t$ when following policy $\pi$.  As explained in  \citep[section
6.2]{Tishby2011:information-to-go}, we assume that the prior of the joint state and action distribution is factorisable, as shown in \eqref{S-eq:state-action-probability-sequence}.  More specifically, that the joint action distribution can be factorised by policy-consistent distributions according to $\Pr(A_t=a_t, \dots, A_T=a_T) = \hat{p}(a_{t})\cdot \hat{p}(a_{t+1}) \cdots \hat{p}(a_{T})$, and that for the joint state distribution,
$\Pr(S_{t+1}=s_{t+1}, \dots,S_T=s_T) = \hat{p}(s_{t+1})\cdot
\hat{p}(s_{t+2}) \cdots \hat{p}(s_{T})$, where the probabilities $\hat{p}(s_{t+1}), \hat{p}(s_{t+2}), \dots$ are time-homogeneous.  The prior state distribution $\hat{p}(s)$ is chosen as a stationary uniform distribution over the states.

\begin{align}
	\bar{\Im}^{\pi}(s_t) = & \E_{\pi(a|s_t)p(s_{t+1}|s_t, a)} \left[ \log{\frac{p(a_t, s_{t+1}|s_t)}{\hat{p}(a_t, s_{t+1})}} + \log \frac {p(a_{t+1},s_{t+2}, a_{t+2}, s_{t+3}\dots|s_{t+1})}{\hat{p}(a_{t+1}, s_{t+2}, a_{t+2},\dots)}\right]\label{eq:information-to-go-sepd}\\
	= &  \sum_{a , s_{t+1}}p(a_t,s_{t+1}|s_t) \biggl[\log{\frac{p(a_t, s_{t+1}|s_t)}{\hat{p}(a_t, s_{t+1})}} \nonumber \\
	& + \sum_{a, s_{t+1}}p(a_{t+1},s_{t+2},\dots|s_{t+1})\log \frac {p(a_{t+1},s_{t+2}, a_{t+3}, \dots|s_{t+1})}{\hat{p}(a_{t+1}, s_{t+2}, a_{t+2},\dots)} \biggr]\\
	= & \E_{\pi(a|s_t)p(s_{t+1}|s_t, a)}\left[\log{\frac{\pi(a_t|s_t)}{\hat{p}(a_t)}} + \log{\frac{p(s_{t+1}|s_t, a_t)}{\hat{p}(s_{t+1})}}+ \Itg_{\pi}(s_{t+1})\right] \label{eq:information-to-go-bellman}
\end{align}
	
Modelling the decision process using the perception-action loop \citep{Tishby2011:information-to-go}, we can use Information-to-go to understand the circular flow of information between the agent and the environment. Information-to-go comprises terms signifying the \emph{environmental response} and the \emph{decision complexity} for each state-action transition  \cite{Tishby2011:information-to-go}.	

The environmental response term, $\E_{\pi(a_t|s_t)p(s_{t+1}|s_t,a_t)}[\log \frac{p(s_{t+1}|s_t,a_t)} {\hat{p}(s_{t+1})}]$, is the information processed by the environment for the state transition, analogous to the information gained if the agent could fully observe the successor state. In a world with a deterministic transition model and a uniform prior state distribution, the term within the expectation reduces to a constant value of $\log \frac{1}{|\S|}$.  In this situation, minimising the environmental response also minimises the number of decisions to reach the goal.  

The environmental response term can also be interpreted as the statistical surprise in the transition due to the selected action.  It reaches a minimum when $p(S_{t+1}|s_t, a_t) = \hat{p}(S_{t+1})$, thus in the case of non-deterministic transitions and with a uniform prior, minimising surprise means that the transition probability is independent from the chosen action and maximises uncertainty regarding the successor state.


\subsection{Decision Information}\label{S-decision-information}

Decision Information is a measure of the surprise of the agent about the future actions in the decision sequence given the action's prior, which in our case is $\hat{p}(a;\pi)$, the marginalised action distribution calculated using the live distribution.  Decision information is therefore  indicative of the information processed by an agent to implement the decision strategy $\pi$.  It is a derivative of Information-to-go, focussing on the decision complexity term $\log \frac{\pi(a_t|s_t)}{\hat{p}(a_t)}$ to the exclusion of the environmental response term.  From the factorisation shown in \eqref{S-eq:state-action-probability-sequence} we define Decision Information as follows:
\begin{align}
	\Itg_D^\pi(s_t) &\defeq \E_{p(a_t, s_{t+1},a_{t+1},\dots|s_t)} \left[ \log \frac{\pi(a_t|s_t)\pi(a_{t+1}|s_{t+1})\dots\pi(a_T|s_T)}{\hat{p}(a_t;\pi)\hat{p}(a_{t+1};\pi)\dots\hat{p}(a_T;\pi)}\right] \\
	&= \E_{p(a_t, s_{t+1},a_{t+1},\dots|s_t)} \left[\log\frac{\pi(a_t|s_t)}{\hat{p}(a_t;\pi)} + \log \frac{\pi(a_{t+1}|s_{t+1})\dots\pi(a_T|s_T)}{\hat{p}(a_{t+1};\pi)\dots\hat{p}(a_T;\pi)} \right] \\
	&= \E_{\pi(a|s_t)p(s_{t+1}|s_t, a)} \left[\log\frac{\pi(a_t|s_t)}{\hat{p}(a_t;\pi)} + \Itg_D^\pi(s_{t+1})\right]  \label{S-eq:itg-d-bellman}
\end{align}

Similar to Information-to-go, Decision Information is a KL divergence and therefore nonnegative, further the Bellman recursive relation also holds for Decision Information \eqref{S-eq:itg-d-bellman}.

\subsection{InfoRL}\label{S-InfoRL}

\citeauthor{Rubin2012} \cite{Rubin2012}, subsequent to the paper on Information-to-go, also ignores the flow of information between the agent and the environment defining InfoRL as the KL divergence between the policy and a fixed default policy $\rho(a|s)$ as follows:
\begin{equation}
	\Delta I(s) = \sum_a \pi(a|s)\log \frac{\pi(a|s)}{\rho(a|s)}.
\end{equation}

\subsection{Discounted Information}\label{S-discounted-information}

\citeauthor{Larsson2017} \citep{Larsson2017} modify InfoRL by defining a discounted information cost $D^\pi$ discounted by factor $\gamma$ for a decision maker starting at state $s$, following policy $\pi(a|s)$ as shown in \eqname \eqname \eqref{eq:discounted-information}.  As in InfoRL, a prior policy $\rho(a|s)$ is selected  and the posterior policy $\pi(a|s)$ is permitted to deviate from the prior according to a threshold value of $D_{KL}[\pi(a|s)|\rho(a|s)]$.  

\begin{equation}
	D^\pi(s_t) \defeq \lim_{T\to\infty}\quad \E_{p(a_t, s_{t+1},a_{t+1},\dots|s_t)}\left[\sum_{t=0}^{T-1}\gamma^t \log \frac{\pi(a_t|s_t)}{\rho(a_t|s_t)}\right]. \label{eq:discounted-information}
\end{equation}

The study also addresses whether it is possible to find an optimal prior $\rho(a|s)$ which minimises the discounted information cost across all states on average by using a state independent prior distribution over actions $\rho(a)$.  Different to our study, \citeauthor{Larsson2017} focus on exploitation by using a prior fixed probability distribution over states to calculate $\rho(a)$ forcing resource limited agents to find a single action distribution that on average yields a high reward across the state space. The study continues to investigate hierarchical state abstractions where admissible action spaces are state dependent.

\subsection{Mutual Information Regularisation}\label{S-mutual-information-regularisation}

The ``one-step entropy regularistion'' form of \textit{Mutual Information Regularisation (MIR)} is analogous to Relevant Information \citep[see section 2.3] {Grau-Moya2019}, which is extended to a sequence of decisions by defining $\mathcal{V_{\pi, \rho}}$ as a value function which incorporates an information cost, see below.  
\begin{equation}
    \mathcal{V}_{\pi,\rho}(s) \defeq \E\left[\sum_{t=0}^\infty \gamma^t \left(r(s_t, a_t) - \frac{1}{\beta} \log\frac{\pi(a_t|s_t)}{\rho(a_t)}\right)\Big\vert s_0 = s\right].
\end{equation}

High entropy policies tend to spread the probability mass across all actions equally and are therefore limited in terms of their exploration of reinforcement learning state spaces.  Thus, the objective of MIR is to dynamically adjust the importance of actions while learning by replacing the uniform prior policy $\rho(a|s)$ with an adaptive distribution of actions $\rho(a)$.  The optimal policy takes the form of a Boltzmann distribution weighted by the prior $\rho$.  The $\beta$ parameter is employed as a penalisation for deviating from the prior which is adapted according to a schedule to favour small values of $\beta$ initially and permitting the agent to explore an extended range of policies as training progresses.

\section{Minimising Free Energy}\label{S-minimising-free-energy}

Constraints on information processing are formulated using a
generalisation of rate-distortion \citep[see chapter 10]{Cover2006},
thus these measures can be calculated via a Blahut-Arimoto-style
algorithm \citep{Blahut1972, Arimoto1972} with alternate iterations of
the constrained policy and an objective function ($Q_\Fen$). Following \citep{Larsson2017}, we compute the policy which minimises free energy, for which we compute Decision Information as described below. To recap, we define free energy as the Lagrangian of the value function constrained by Decision Information, as detailed below and described in the main paper. 
\begin{equation}
		\Fen^{\pi}(s;\beta) \defeq \frac{1}{\beta}\Itg^{\pi}_{D}(s) - V^{\pi}(s).\label{S-eq:free-energy}
	\end{equation}

\subsection{Computing Constrained Free Energy}\label{S-computing-itg}
Minimising the free energy \eqref{S-eq:free-energy} directly enables us to find the policy $\pif$ which achieves the best performance under a constraint on information processing specified by the trade-off parameter $\beta$, $\pif = \argmin_{\pi(a|s)}\Fen^\pi(s; \beta)$.
Similarly to \citet{Rubin2012} and further clarified by \citet{Larsson2017} we proceed as described below.  For convenience, we adopt the notation $s=s_t$, $a=a_t$ and $s' = s_{t+1}$. Below we show the derivation of the Lagrange method to solve the constrained optimisation, which minimises free energy using Lagrange multipliers: $\beta$, which behaves as a trade-off parameter, and $\lambda$ for the normalisation constraint for a valid probability distribution. 

{\footnotesize
\begin{align}
	\L(s;\pi, \beta, \lambda) &\defeq \frac{1}{\beta}\Itg_{D}^{\pi}(s) - V^{\pi}(s) + \lambda\left[1-\sum_{a}\pi(a|s)\right] \\
    &= \sum_{s', a}{p(a, s'|s)}\left[\frac{1}{\beta}\log{\frac{\pi(a|s)}{\hat{p}(a; \pi)}} + \frac{1}{\beta}\Itg^{\pi}_D(s') - r(s', a, s) - V^{\pi}(s')\right] + \lambda\left[1-\sum_{a}\pi(a|s)\right]\\
    &= \sum_{a}\pi(a|s)\sum_{s'}p(s'|s,a)\left[\frac{1}{\beta}\log{\frac{\pi(a|s)}{\hat{p}(a; \pi)}} - r(s', a, s) + \Fen^\pi(s';\beta) \right] + \lambda\left[1-\sum_{a}\pi(a|s)\right].
\end{align}
For convenience we define 
\small{\begin{align}
    Q^\pi_\Fen (s, a ; \beta) &\defeq \sum_{s'}p(s'|s,a)\left( r(s', s, a) - \Fen^{\pi}(s'; \beta)\right) \label{eq:Qf} \quad  \text{ then }\\
    \L(s; \pi, \beta, \lambda) &= \sum_{a}\pi(a|s)\sum_{s'}p(s'|s,a)\left[\frac{1}{\beta}\log{\frac{\pi(a|s)}{\hat{p}(a; \pi)}}\right] - \sum_a\pi(a|s)Q_\Fen^\pi(s, a ; \beta) +\lambda\left[1-\sum_{a}\pi(a|s)\right].
   \end{align}

To solve for the policy which minimises the free energy we take the partial derivative of the Lagrangian with respect to the policy and set it to zero. We assume that $\Fen^\pi$ and therefore $Q^\pi_{\Fen}$ change slowly with respect to $\pi$ in comparison to the other terms of the Lagrangian and we therefore treat it as constant with respect to the policy. For convenience when taking derivatives, we treat binary logarithms as natural logarithms as the change of base terms cancel out. For the derivative of a particular policy we choose $\pi$ with given $\bar{a}$ and $s$, i.e.  $\pi(\bar{a}|s)$, which simplifies with the summation over $a$. 

\begin{align*}
    0&=\frac{\partial }{\partial \pi(\bar{a}|s)}\L(s;\pi, \beta, \lambda) \\
    &= \frac{\partial }{\partial \pi(\bar{a}|s)} \left( \sum_{a}\pi(a|s)\sum_{s'}p(s'|s,a)\left[\frac{1}{\beta}\log{\frac{\pi(a|s)}{\hat{p}(a; \pi)}}\right] - \sum_a\pi(a|s)Q_\Fen^\pi +\lambda\left[1-\sum_{a}\pi(a|s)\right]\right)\\
    &=\frac{\partial }{\partial \pi(\bar{a}|s)} \left( \pi(\bar{a}|s)\sum_{s'}p(s'|s,\bar{a})\left[\frac{1}{\beta}\log{\frac{\pi(\bar{a}|s)}{\hat{p}(\bar{a}; \pi)}}\right] - \pi(\bar{a}|s)Q_\Fen^\pi +\lambda\left[1-\pi(\bar{a}|s)\right] \right) \\
    &=\left( \log{\frac{\pi(\bar{a}|s)}{\hat{p}(\bar{a}; \pi)}} + \frac{\pi(\bar{a}|s)}{\pi(\bar{a}|s)}  \right) \sum_{s'}p(s'|s,\bar{a})-\beta Q_\Fen^\pi - \beta \lambda  \\
\log\pi(\bar{a}|s) &= \beta Q_\Fen^\pi +\beta \lambda  -1 + \log\hat{p}(\bar{a};\pi) \\
\pi(\bar{a}|s) &= \hat{p}(\bar{a};\pi)\cdot\exp(\beta Q_\Fen^\pi)\cdot\exp(\beta \lambda)\cdot\exp(-1)
\end{align*}
}%
If we don't assume natural logarithms then the term $\exp(-1)$ becomes $\exp\frac{-1}{\ln 2}$.  For a valid probability distribution $\sum_{a}\pi(a|s) = 1$, therefore

\begin{align*}
    \sum_a\pi(a|s) &= \sum_a \hat{p}(a;\pi)\cdot\exp(\beta Q_\Fen^\pi)\cdot\exp(\beta \lambda)\cdot\exp(-1) = 1\\
    \exp(\beta \lambda) &= \frac{1}{\sum_a \hat{p}(a;\pi)\cdot\exp(\beta Q_\Fen^\pi)\cdot\exp(-1)} \text{ then }\\
	\pi(\bar{a}|s) &= \frac{\hat{p}(\bar{a};\pi)\cdot\exp(\beta Q_\Fen^\pi)}{\sum_{a}\hat{p}(a;\pi)\exp(\beta Q_\Fen^\pi)}
\end{align*}
The partition function $Z$ which is later used for normalisation of the policy is then defined in terms of the marginal of the policy as
\begin{equation}
    Z(s; \beta) \defeq \sum_{a} \hat{p}(a;\pi)\exp \left[\beta Q_\Fen(s,a;\beta)\right], \label{eq:itg-Z}
\end{equation} 

with the marginal action distribution
\begin{equation}
    \hat{p}(a; \pi) = \sum_{s \in \mathcal{S}}\pi(a|s)\hat{p}(s; \pi). \label{S-eq:marginal-action-distribution}
  \end{equation}

We then find an approximation to the free energy by substituting the equation for the policy \eqref{eq:itg-policy} back into the original Lagrangian equation, the term in $\lambda$ is zero as $\sum_a\pi(a|s)=1$, hence we can remove the renormalisation constraint, which is simply the free energy.

\begin{equation}
	\pi(\bar{a}|s) = \frac{\hat{p}(\bar{a}; \pi)} {Z(s; \beta)} \exp \left[\beta Q_\Fen(s,\bar{a};\beta)\right]\label{eq:itg-policy}.
  \end{equation} 

\begin{align}
	\Fen(s;\beta)  \approx &\E_{\pi(a|s)p(s'|s,a)}\left[\frac{1}{\beta}\log\frac{\pi(a|s)}{\hat{p}(a;\pi)} - r(s',s,a) + \Fen^\pi(s';\beta)\right]\\
    \begin{split}
    = & \sum_{a}\frac{\hat{p}(a;\pi)}{Z(s;\beta)}\exp(\beta Q_\Fen^\pi(s, a;\beta))\sum_{s'}p(s'|s,a)\\
	&\left[\frac{1}{\beta}\log{\frac{\frac{\hat{p}(a;\pi)}{Z(s;\beta)}\exp(\beta Q_\Fen^\pi(s, a;\beta))}{\hat{p}(a; \pi)}} - r(s', a, s) + \Fen^\pi(s';\beta) \right]
    \end{split}\\
	\begin{split}
	=&\sum_{a}\frac{\hat{p}(a;\pi)}{Z(s;\beta)}\exp(\beta Q_\Fen^\pi(s, a;\beta)) \\
	&\sum_{s'}p(s'|s,a)\frac{1}{\beta}\left[\log\frac{1}{Z(s;\beta)}+\beta Q_\Fen^\pi(s,a;\beta))\right] -Q_\Fen^\pi(s,a;\beta) \end{split}\\
    &= -\frac{1}{\beta}\log Z(s;\beta)\label{eq:free-energy-Z}
\end{align}

This is consistent with the well known fact that the Boltzmann distribution solves the corresponding free energy \citep[see section 33.1]{Mackay2003}, which can thus be approximated using the partition function (\eqref{eq:free-energy-Z}).

For a fixed $\beta \in (0, \infty)$, we have computed the free energy optimal policy in terms of $\hat{p}(a;\pi)$, which itself depends on $\pi(a|s)$.  Thus we iterate to find a solution (as detailed in algorithm \ref{S-alg:policy-free-energy} below).  Starting with a random guess for $\pi(a|s)$, we update the prior state and action distributions, and then calculate the partition function, policy and free energy until the policy and free energy converge.  For the policy, convergence is determined whether or not the Kullback-Leibler divergence between consecutive policies is less than $\varepsilon_\pi$, and for free energy, if the difference betweeen consecutive free energy values is less than $\varepsilon_\Fen$.  In this study both $\varepsilon_\pi = 1\times 10^-5$ and $\varepsilon_\Fen = 1\times 10^-5$. The assumption that $Q^\pi_\Fen(s, a; \beta)$ is constant with respect to the policy when solving the Lagrangian is acceptable because the policy and free energy solutions are alternated in the double iteration of the Blahut-Arimoto-style algorithm. The corresponding Decision Information can then be calculated using the fixed-point equation \eqname \eqref{eq:itg-d-bellman} and iterated until convergence.

\vspace{1em}

\begin{algorithm}[ht!]
	\caption{Given $\beta$, find free energy optimal policy $\pi(a|s)$ and calculate $\Fen^{\pi}(s; \beta)$.}
	\label{S-alg:policy-free-energy}
	\begin{algorithmic}[0]
		\Require transition dynamics $p(s'|s, a)$, reward function $r(s', s, a)$,  trade-off parameter $\beta$
		\State Actions: $\mathcal{A} = \{a_1, \dots, a_m\}$
		\State States: $\mathcal{S} = \{s_1, \dots, s_n\}$
		\State Initialise: Free-energy function: $\Fen \gets \mathbf{0}$
		\State Initialise: Policy to be uniformly distributed, $\pi(a|s)  \thicksim $$ \mathbf{\mathcal{U}}:\frac{1}{|\A|}$
		\Repeat
			\State $\Fen' \gets \Fen$
			\State $\pi' \gets \pi$
			\State update $\hat{p}(s; \pi)$ given $\pi(a|s)$ \Comment{live distribution, section \ref{S-live-state-distribution}}
			\State $\hat{p}(a;\pi) \gets \sum_{s} \pi(a|s)\hat{p}(s; \pi)$ \Comment {marginal action distribution \eqref{S-eq:marginal-action-distribution}}
			\ForAll {$s \in \S$}
				\ForAll {$~a \in \A$}
					\State $Q_\Fen (s, a ; \beta) \gets \sum_{s'}p(s'|s,a)\left( r(s', s, a) - \Fen'(s'; \pi, \beta)\right)$ \Comment{\eqref{eq:Qf}}
				\EndFor
				\State $Z(s;\beta) \gets \sum_{a}\hat{p}(a;\pi)\exp \left[\beta Q_{\Fen}(s, a;\beta)\right]$ \Comment{\eqref{eq:itg-Z}}
				\State $\Fen(s; \beta) \gets -\frac{1}{\beta} \log Z(s; \beta)$ \Comment{\eqref{eq:free-energy-Z}}
				\ForAll {$~a \in \A$}
					\State $\pi(a|s) \gets \frac{\hat{p}(a;\pi)} {Z(s; \beta)} \exp \left[\beta Q_\Fen(s,a;\beta)\right]$ \Comment{\eqref{eq:itg-policy}}
				\EndFor
			\EndFor
		\Until{Convergence:$\vert \Fen - \Fen^{\prime}\vert \leq \varepsilon_{\Fen}$ and $D_{KL}[\pi\Vert\pi ']\leq \varepsilon_{\pi}$}
	\end{algorithmic}
	\end{algorithm}
	
\subsection{Computational Complexity}\label{S-computational-complexity}

We computed the free energy using Algorithm 1, which is reported above. This is an iterative algorithm that computes subsequent estimates of the free energy until these converge toward a fixed point. The computational complexity of Algorithm 1 is $\mathcal{O}(K \lvert \S\rvert^2 ( \lvert A\rvert + \lvert \S \rvert^{0.4}))$.  Here, the $\mathcal{O}(\lvert \S \rvert^{2.4})$ term represents the cost of inverting the state transition matrix needed to compute the live distribution (when using the Coppersmith–Winograd algorithm). Furthermore, $K$ denotes the number of iterations necessary to reach convergence. Although the exact estimation of $K$ is still an open research question, it can be indicative to mention the rates of convergence of two closely related iterative algorithms: the Blauth Arimoto (BA) and Value Iteration (VI) algorithms. In fact, Algorithm 1 can be seen as a combination of the BA algorithm, which is used to solve rate-distortion problems in information theory, and the VI algorithm, which is used to compute the optimal policy of a MDP. The computational complexity of the BA algorithm is $\mathcal{O}(K_{BA}M N^2)$, where $N$ denotes the input alphabet size, $M$ denotes the alphabet size of the source code, $\mathcal{O}(M N^2)$	 is the computational complexity of one iteration and $K_{BA} = \frac{1}{\varepsilon} \log{}N$ is the number of iterations needed to find an $\varepsilon-$approximation of the rate-distortion function \citep{sutter2014efficient}. The computational complexity of VI is $\mathcal{O}(K_{VI} \lvert \S \rvert^2 \lvert A \rvert)$, where $\mathcal{O}(\lvert \S \rvert^2 \lvert A \rvert)$ is the computational complexity of one iteration and $K_{VI}$ denotes the number iterations needed for convergence, which is a polynomial function of $\lvert \S \rvert$ and $\lvert A \rvert$ \citep{littman1995complexity}.

In order to find all the $\varepsilon-$infodesics of a MDP, first the free energies between every couple of states need to be computed. These are used as weights of the weighted directed graph $(\S, E)$ that represents our cognitive geometry. This requires at most $\lvert \S \rvert(\lvert \S \rvert-1) = \mathcal{O}(\lvert \S \rvert^2)$ free energy computations.  Then, to find all the sequences of states that form $\varepsilon-$infodesics, this graph needs to be searched to find all the paths from state $s$ to state $g$ of weighted length smaller than $\Fen^\pif_g(s)(1+\varepsilon)$ for all $s$ and $g$. This can be done using a computational time of $\mathcal{O}(\lvert \S \rvert(\lvert E \rvert + \lvert \S \rvert \log{}  \lvert \S \rvert + p \lvert \S \rvert))$ (adopting the algorithm reported in \citep{eppstein1998finding}), where $p$ is the number of states of the $\varepsilon-$infodesics with the longest sequence. Hence, the total computational complexity of finding all the $\varepsilon-$infodesics in a MDP is $\mathcal{O}(K \lvert \S \rvert^4 ( \lvert A \rvert + \lvert \S \rvert^{0.4}) + p \lvert \S \rvert^2)$, which can be quite expensive for large MDPs or high $\varepsilon$.

\section{Trade-off between Performance and Information Processing Costs}

Because agents are often required to act parsimoniously when processing information \citep{Polani2009, Genewein2015, Marzen2017, Lieder2020:resource-rational-cognition}, we wish to understand the consequences of these constraints.  In minimising free energy, the trade-off parameter $\beta$ \eqref{eq:free-energy} determines how parsimonious an agent is with respect to information processing.  As $\beta$ decreases, the optimal policy with respect to free energy processes increasingly less information, indicated by lower $\ItgDpif(s)$ values. In the limit of $\beta \to 0$, the agent is moving towards an open-loop policy which processes no information, therefore $\ItgDpif \to 0$ and the policy approaches the marginal $\hat{p}(a;\pif)$. 

 As $\beta$ increases, minimising free energy cares less about information and more about performance, thus the resulting optimal policy requires more information to be processed.  Executing the policy thus makes increasing use of information until, increasing $\beta$ further, the performance approaches the reward-optimal value $V^*(s)$ and the policy approaches $\piv$, where the information processing cost reaches its maximum and the agent takes a shortest path ($\beta > 100$ in our experiments).

This trade-off between cognitive burden and performance \citep{Dijk2013} is shown in  \figurename \ref{figS1:trade-off-curves} by plotting behavioural performance measured by the average value function ($\E_{S\sim 1/{\vert \S \vert}}[V^\pi(s)]$) against the average Decision Information ($\E_{S\sim 1/{\vert \S \vert} }[\Itg^\pi_D(s)]$).   Manhattan and Moore neighbourhoods are represented on the graph by solid and unfilled markers respectively, the shape of the marker is indicative of the goal location.

\marginpar{
\vspace{.7cm} 
\small
\color{captionColour} 
\textbf{Figure \ref{figS1:trade-off-curves}. Trade-off curves} 
for Manhattan ($\shortarrow{0}, \shortarrow{2}, \shortarrow{4}, \shortarrow{6}$) and Moore ($\shortarrow{0},  \shortarrow{1}, \shortarrow{2},  \shortarrow{3}, \shortarrow{4},  \shortarrow{5}, \shortarrow{6}  \shortarrow{7}$) neighbourhoods in a deterministic 5$\times$5 gridworld. Three goal states are considered: the middle, the corner of the world, and on the diagonal between the centre and the corner.   Expected performace ($\E[V]$) is plotted against expected information ($\E[\Itg^\pi_D]$) for various $\beta$ values ranging from $1\times 10^{-3}$ to $100$. 
}
\begin{figure}[!ht]
\includegraphics[trim={0.1cm 2 0.5 0.5}, clip, width=120mm]{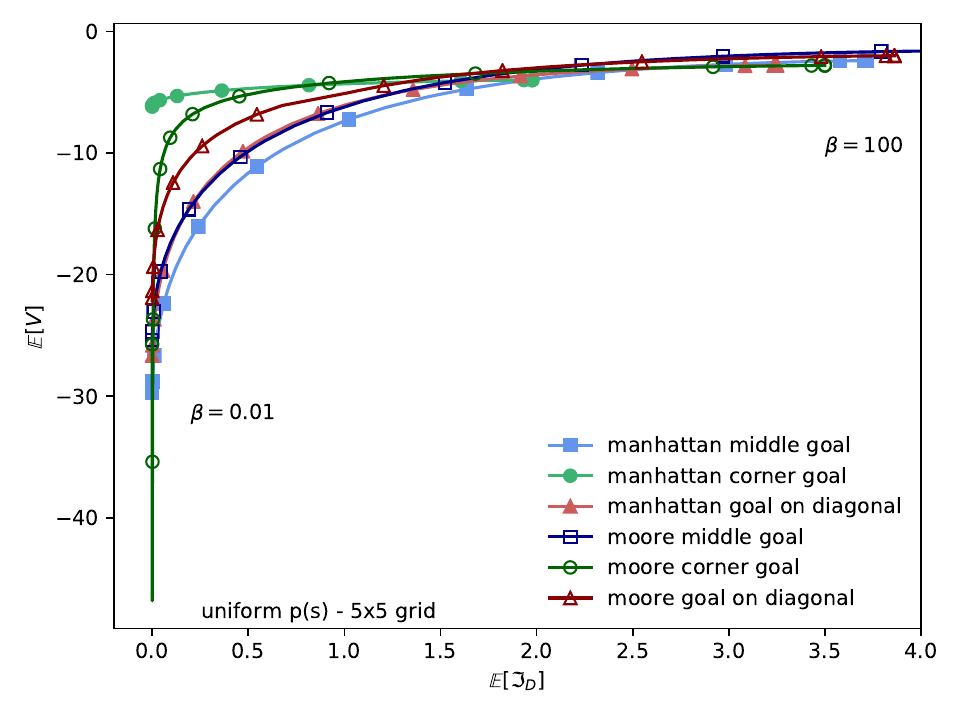}
\captionsetup{labelformat=empty} 
\caption{} 
\label{figS1:trade-off-curves} 
\end{figure} 

Because an ordered set of actions is available in each state \citep{Polani2011}, corner goals are informationally cheaper to reach. In the Moore neighbourhood, an agent can take repeated actions diagonally, where in the case of the Manhattan neighbourhood, the agent randomly alternates between a pair of actions to move approximately diagonally.  In the latter behaviour the agent gets additional guidance along the edges, bumping into them incurs only a minor time penalty. Thus the wall acts as a guide or ``funnel''  buffering some wrong actions on the way to the corner goal.  In contrast, to reach the middle goal, any wrong action is likely to move the agent away from the goal. Thus much more accurate control is required to reach precisely the middle goal compared to drifting into a corner.

In our case, processing less information (low $\beta$ values) leads to the agent being less discerning of its state and thus acting more generally, resulting in longer routes to the goal, i.e.\  lower values for the value function. From the graph it is clear that permitting maximal information processing permits maximising performance, and attaining the optimal value. 

\section{Free Energy Symmetries}\label{S-symmetrised-free-energy}
 Figures \ref{figS2:frees-man-b-100} and \ref{figS3:frees-man-b-0.1} show the free energy values and optimal policies for all combinations of $s,g$ in $\S \times \S$ for $\beta = 100$ and $\beta = 0.1$ respectively.  Using these plots we can observe that the free energy values are not symmetrical when information processing constraints are introduced.

 Figures \ref{figS4:frees-asym-prop-man-b-0.1} and \ref{figS5:frees-asym-prop-man-b-0.01} show the discrepancy between the symmetrised and non-symmetrised free energy values for $\beta$ values of 0.1 and 0.01 respectively as a proportion of the symmetrised values.  The proportion is calculated using the adjacency matrix $D$ and the symmetrised $D^{\mathrm{sym}} \defeq \left(D + D^{\mathop{\intercal}}\right)/{2}$ via $\left( D-D^{\mathrm{sym}} \right)/D^{\mathrm{sym}}$. We observe negligible asymmetry for $\beta \geq 100$.  From the plots presented we observe that the asymmetrical component varies with $\beta$ in the regions between the centre and the corner states.   These plots allow one to estimate the inaccuracies introduced in the visualisation of \figurename \ref{fig3:mds-moo-11-11} owing to this symmetrisation. It is possible to see that, the discrepancies introduced by the symmetrisation are consistent with the topological location of the states in the grid (e.g., all corners' distances are changed  in the same way). Hence, the qualitative distortion of the distance between states introduced by the addition of information processing constraints reported in \figurename \ref{fig3:mds-moo-11-11} is preserved.

 \clearpage
\vspace{.5cm} 
\begin{figure}[!ht]
	\begin{adjustwidth}{-2.0in}{0in}
		\includegraphics[width=\linewidth, trim= 0 -10 0 20, clip]{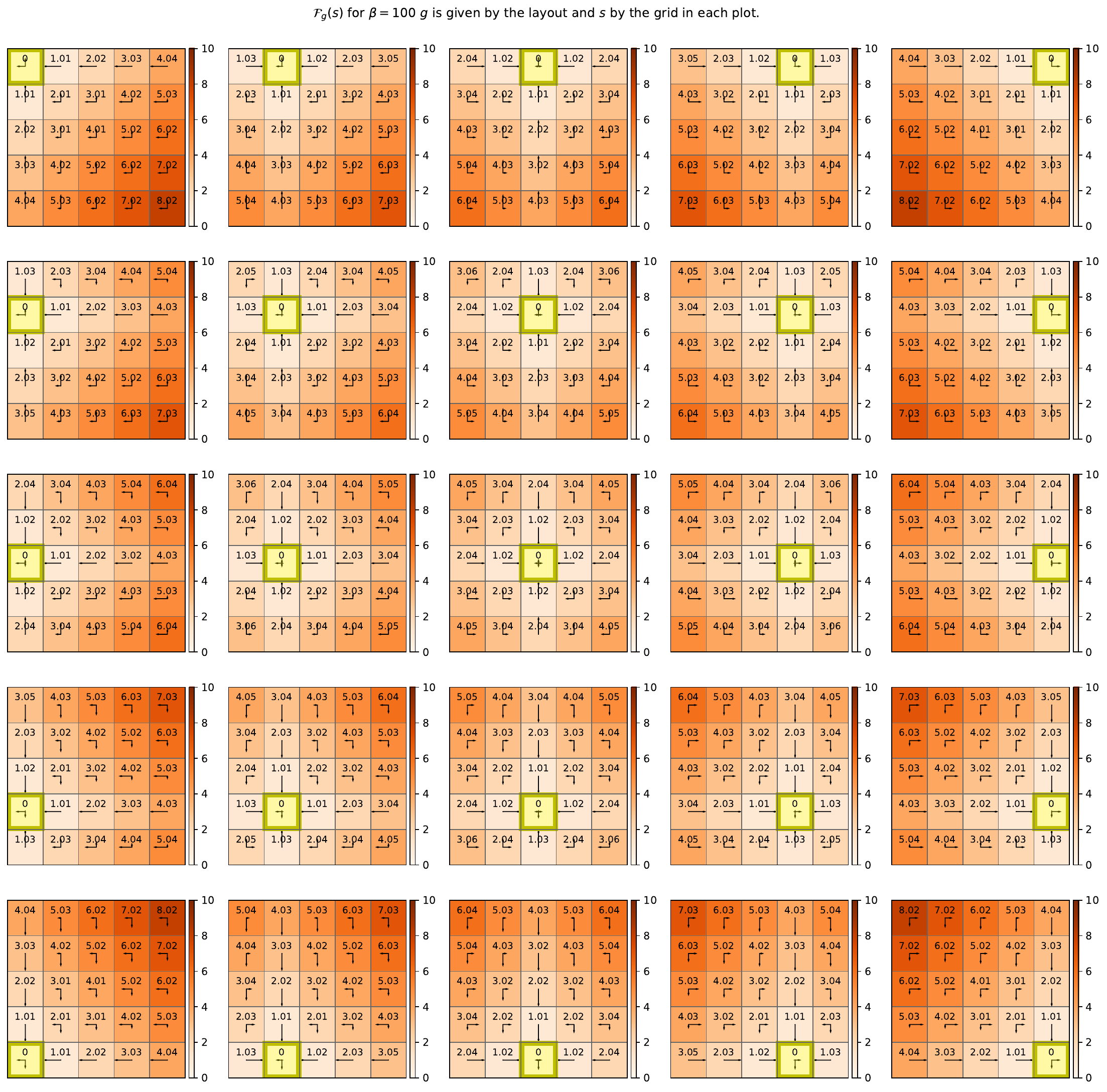}
		\captionsetup{labelformat=empty} 
		\caption{} 
		\label{figS2:frees-man-b-100}
	\end{adjustwidth}
\end{figure}
\vspace{-0.9cm}
\begin{adjustwidth}{-2.0in}{0in}
	\small
	\color{captionColour}
	\textbf{Figure \ref{figS2:frees-man-b-100}. A grid of free energy plots} for a  5 $\times$ 5 gridworld  with cardinal (Manhattan) actions $\A:\{\shortarrow{0}, \shortarrow{2}, \shortarrow{4}, \shortarrow{6}\}$ with $\beta=100$.  The layout of the individual free energy plots in a grid is indicative of the goal state $g$, e.g. the top left plot has the goal in the top left corner, as confirmed by the zero free energy value.  The arrow lengths are proportional to the conditional probability $\pi(a|s)$ in the indicated direction.  The policy presented is optimal with respect to free energy, i.e $\pif = \argmin_\pi \Fen_g^\pi (s;\beta)$ for all $s \in \S$.  The relevant prior, i.e. the joint state and action distribution marginalised over all transient states, $\hat{p}(a;\pi)$ is shown in the goal state.  The heatmap and annotations show the free energy for each state $\Fen^\pif_g(s)$. 
	\vspace{.5cm}
\end{adjustwidth}

\clearpage
\vspace{.5cm} 
\begin{figure}[!ht]
	\begin{adjustwidth}{-2.0in}{0in}
		\includegraphics[width=\linewidth, trim= 0 -10 0 20, clip]{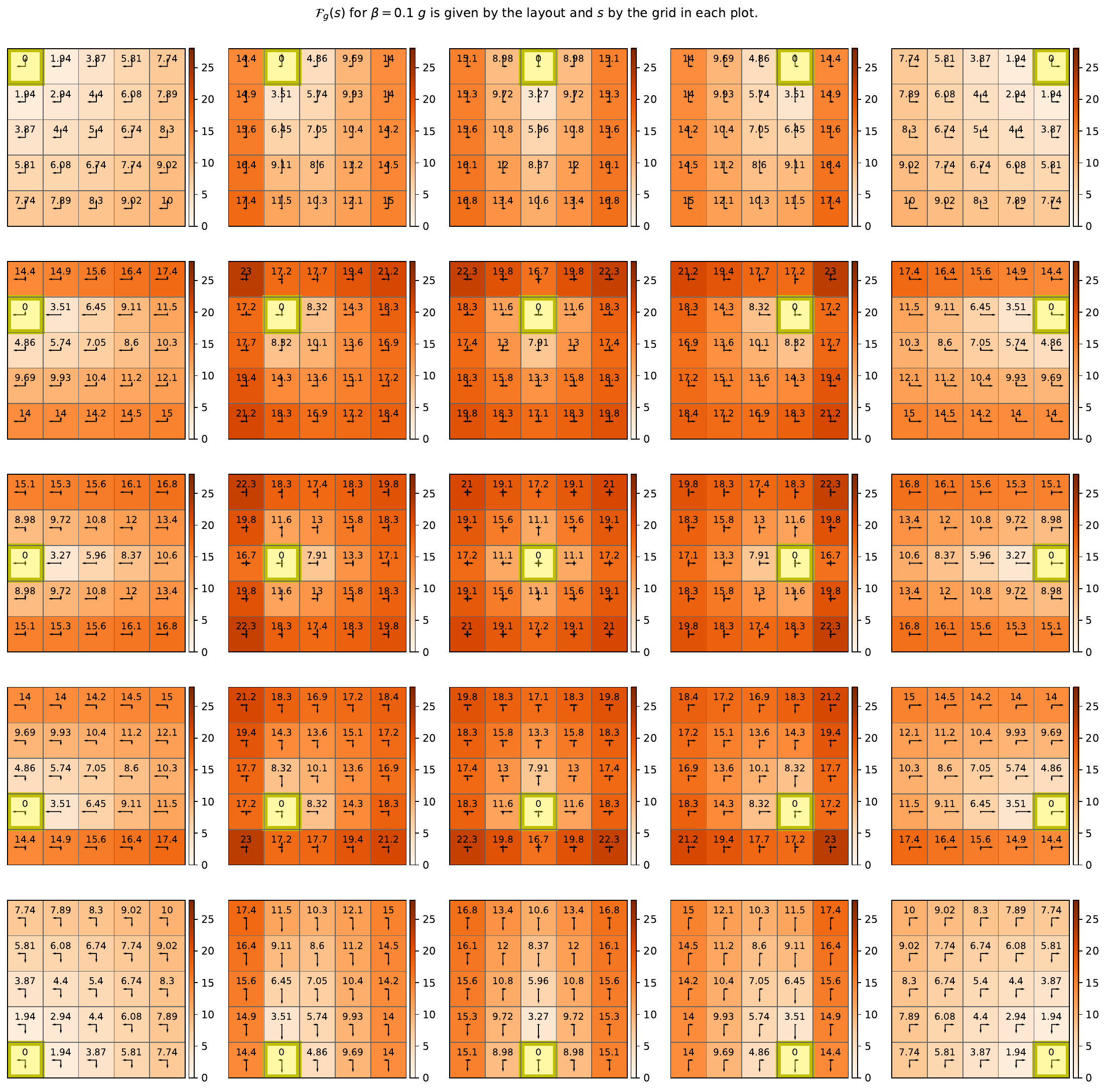}
		\captionsetup{labelformat=empty} 
		\caption{} 
		\label{figS3:frees-man-b-0.1}
	\end{adjustwidth}
\end{figure}
\vspace{-0.9cm}
\begin{adjustwidth}{-2.0in}{0in}
	\small
	\color{captionColour}
	\textbf{Figure \ref{figS3:frees-man-b-0.1}. A grid of free energy plots} for a  5 $\times$ 5 gridworld  with cardinal (Manhattan) actions $\A:\{\shortarrow{0}, \shortarrow{2}, \shortarrow{4}, \shortarrow{6}\}$ with $\beta=0.1$.  The layout of the individual free energy plots in a grid is indicative of the goal state $g$, e.g. the top left plot has the goal in the top left corner, as confirmed by the zero free energy value.  The arrow lengths are proportional to the conditional probability $\pi(a|s)$ in the indicated direction.  The policy presented is optimal with respect to free energy, i.e $\pif = \argmin_\pi \Fen_g^\pi (s;\beta)$ for all $s \in \S$.  The relevant prior, i.e. the joint state and action distribution marginalised over all transient states, $\hat{p}(a;\pi)$ is shown in the goal state.  The heatmap and annotations show the free energy for each state $\Fen^\pif_g(s)$. 
	\vspace{.5cm}
\end{adjustwidth}

\clearpage

\vspace{.5cm} 
\begin{figure}[!ht]
	\begin{adjustwidth}{-2.0in}{0in}
		\includegraphics[width=\linewidth, trim= 0 -10 0 20, clip]{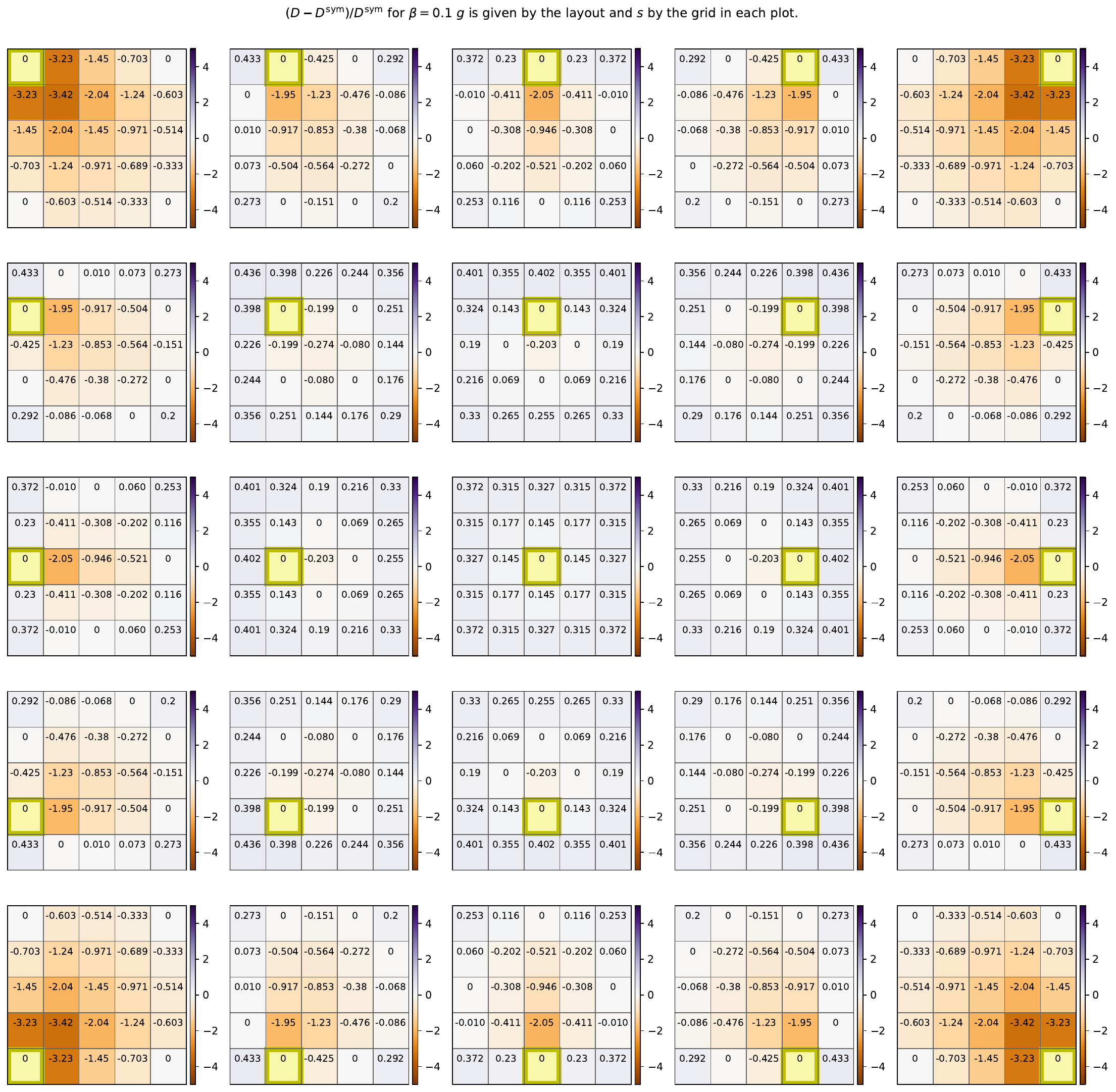}
		\captionsetup{labelformat=empty} 
		\caption{} 
		\label{figS4:frees-asym-prop-man-b-0.1}
	\end{adjustwidth}
\end{figure}
\vspace{-0.9cm}
\begin{adjustwidth}{-2.0in}{0in}
	\small
	\color{captionColour}
	\textbf{Figure \ref{figS4:frees-asym-prop-man-b-0.1}. A grid of free energy symmetry plots} for a  5 $\times$ 5 gridworld  with cardinal (Manhattan) actions $\A:\{\shortarrow{0}, \shortarrow{2}, \shortarrow{4}, \shortarrow{6}\}$ with $\beta=0.1$.  The layout of the individual free energy plots in a grid is indicative of the goal state $g$, e.g. the top left plot has the goal in the top left corner, as confirmed by the yellow outline.   The heatmap and annotations show the proportion of free energy asymmetry for each state given by $( D - D^{\mathrm{sym}})/D^{\mathrm{sym}}$.
	\vspace{.5cm}
\end{adjustwidth}

\clearpage
\vspace{.5cm} 
\begin{figure}[!ht]
	\begin{adjustwidth}{-2.0in}{0in}
		\includegraphics[width=\linewidth, trim= 0 -10 0 20, clip]{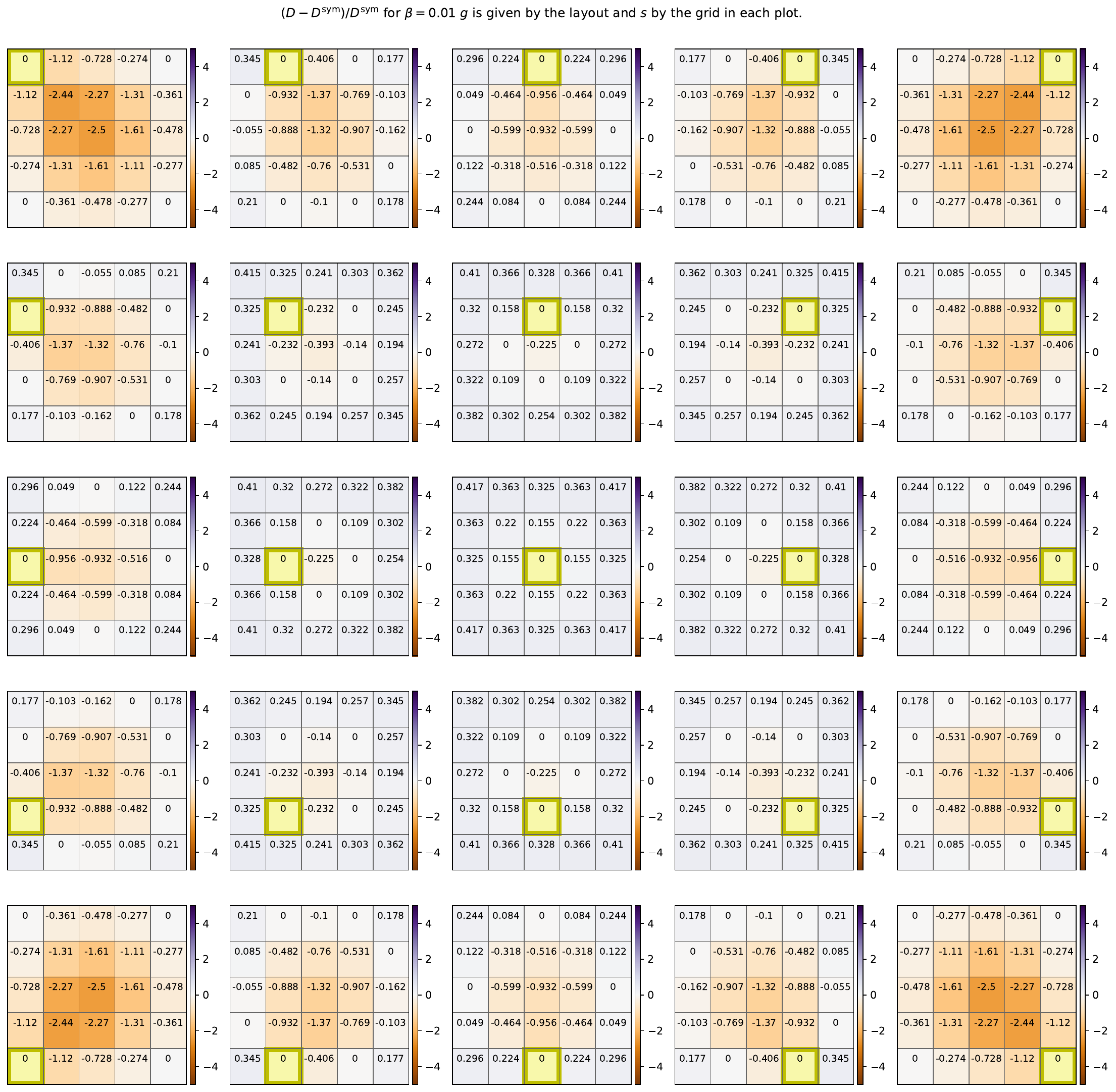}
		\captionsetup{labelformat=empty} 
		\caption{} 
		\label{figS5:frees-asym-prop-man-b-0.01}
	\end{adjustwidth}
\end{figure}
\vspace{-0.9cm}
\begin{adjustwidth}{-2.0in}{0in}
	\small
	\color{captionColour}
	\textbf{Figure \ref{figS5:frees-asym-prop-man-b-0.01}. A grid of free energy symmetry plots} for a  5 $\times$ 5 gridworld  with cardinal (Manhattan) actions $\A:\{\shortarrow{0}, \shortarrow{2}, \shortarrow{4}, \shortarrow{6}\}$ with $\beta=0.01$.  The layout of the individual free energy plots in a grid is indicative of the goal state $g$, e.g. the top left plot has the goal in the top left corner, as confirmed by the yellow outline.   The heatmap and annotations show the proportion of free energy asymmetry for each state given by $( D - D^{\mathrm{sym}})/D^{\mathrm{sym}}$.
	\vspace{.5cm}
\end{adjustwidth}

\clearpage
\section{Triangle Inequality for Disjoint Subpolicies}\label{S-triangle-inequality-disjoint-policies}
We show here that the free energy in goal-directed
navigation satisfies the triangular inequality if the trajectories
produced by the subpolicies for intermediate goals never overlap
(except at the subgoals themselves). Hence, in this case the free energy is a quasi-metric on the state space $\S$.
We begin with several definitions.
In the main paper we have introduced the optimal policy achieving the
minimum free energy $\pif$ which is defined as follows:

\begin{definition}
Given a goal state $g \in \S$, define $\pif$ via $\pif(s) \defeq \arg\min_{\pi} \Fen^{\pi}_g(s) $ for all $s \in \S$.
\end{definition}

For this section it will be useful to define also the set of all states visited by an agent using a policy. 

\begin{definition}
  Given a policy $\pi$ and a state $s \in \S$, define $\S_{\pi(s)}$ as
  the set containing all the states eventually visited by an agent
  with probability larger than 0 when starting in state $s$ and then
  following the policy $\pi$. We include the starting state in
  $\S_{\pi(s)}$, i.e.\ $s\in \S_{\pi(s)}$.
\end{definition}

In the Discussion, we have seen that, to form a proper quasi-metric on the state space $\S$, the free energy $\mathcal{F}^{\pif}_g(s)$ from $s$ to $g$ must satisfy the following triangular inequality 

\begin{equation}
\Fen^{\pif}_g(s) \leq \Fen^\pifi{1}_{s'}(s) + \Fen^\pifi{2}_g(s'),
\label{eq:inequality}
\end{equation} 

where $s$ is the agent's starting state,  $g$ is the goal state and
$s'$ any interim subgoal.  The optimal policy $\pif$ directly brings
the agent from $s$ to $g$ without constraining possible interim
states. In contrast, when composing the optimal subpolicies
$\pifi{1}$ and $\pifi{2}$, the agent firstly moves from $s$ to $s'$
(using $\pifi{1}$) and then moves from $s'$ to $g$ (via $\pifi{2}$).
We remind the reader that underlying the individual free energy terms
in (\ref{eq:inequality}) there are actually distinct MDPs (in our
formalism goals are modelled as absorbing states, see main text for
details); we then hierarchically combine the two policies when considering the composition of the two terms in the right hand side of the inequality. We have already shown with a counterexample that the triangular inequality (\ref{eq:inequality}), in its general form,  does not hold (main text, \eqname\eqref{eq:infodesic-relaxed}).  In this section, we prove the triangular inequality when the subpolicies are \emph{essentially disjoint}, i.e., if starting at $s$, the trajectories induced by $\pifi{1}$ and those by $\pifi{2}$ only intersect trivially at $s'$ (namely, in the case of $\pifi{1}$ as goal of the first policy, in the case of $\pifi{2}$ as starting state of the second policy.) 

We proceed now by proving the following theorem:

\begin{theorem}
  Let $s \neq g$, $s,g \in \S$ be given. Then, if for all $s'\in \S$ we have
  $\S_{\pifi{1}(s)} \bigcap \S_{\pifi{2}(s')} = \{ s' \}$, then
  $\Fen^{\pif}_g(s) \leq
  \Fen^{\pifi{1}}_{s'}(s) + \Fen^{\pifi{2}}_g(s')$
\end{theorem}

\textbf{Proof} For $s' = g$ the theorem can be easily proven, since we
have $\Fen_g(g) = 0$, then $\Fen^{\pif}_g(s) =
\Fen^{\pifi{1}}_{s'}(s) + 0 = \Fen^{\pifi{1}}_{g}(s)$, which fulfils
the triangular inequality, and analogously for $s'=s$.  Therefore,
assume    $s \neq s' \neq g$ and we employ proof by contradiction. We will show that, if there exists a $s'\neq g$ violating this condition (i.e. for which the inequality is reversed and strict), then we can glue the subpolicies $\pifi{1}$ and $\pifi{2}$ together in a way such that the resulting new policy $\hat{\pi}_\Fen$ achieves a smaller overall free energy than the original optimal policy $\pif$, leading to a contradiction.

Formally, assume that  there exists $s' \in \S$, with $s' \neq g$, for which $\Fen^{\pif}_g(s) > \Fen^{\pifi{1}}_{s'}(s) + \Fen^{\pifi{2}}_g(s')$. Then, construct a new policy $\hat{\pi}_\Fen$, via:

\[
  \forall q \in \S: \;\;\; \hat{\pif}(a|q) \defeq
  \begin{cases}
                                  \pifi{1}(a|q) & \text{if $q\in \S_{\pifi{1}(s)} \setminus \{ s' \}$} \\
                                   \pifi{2}(a|q) & \text{if $q \in \S_{\pifi{2}(s')}$}
  \end{cases}
\]

Note that, similar to $\pif$, also $\hat{\pi}_\Fen$ can be used by an
agent to reach the goal $g$ from the initial state $s$. Furthermore,
since we assumed $\S_{\pifi{1}(s)}  \bigcap \S_{\pifi{2}(s')} = \{ s'
\}$,  $\hat{\pi}_\Fen$ is well defined for each state on the possible
trajectories, with no ambiguity as to which subpolicy is to be used.
We can then compute the free energy $\Fen^{\hat{\pi}_\Fen}_g(s)$ as
follows (superscripts indicate whether states and actions belong to
the first or second section of the trajectory: 

\begin{equation}
\begin{aligned}
& \Fen^{\hat{\pi}_\Fen}_g(s) =  \sum_{(a_0^1, s_1^1, a_1^1, \dots, a_0^2, s_1^2, a_1^2, \dots )} p(a_0^1, s_1^1, a_1^1, \dots, s', a_0^2, s_1^2, a_1^2, \dots, g|s) \cdot  \\ 
& \cdot \Big( \frac{1}{\beta} \log \frac{\hat{\pi}_\Fen(a_0^1|s)}{\widehat{p}(a_0^1)} - r(s_1^1,s,a_0^1) +\frac{1}{\beta} \log \frac{\hat{\pi}_\Fen(a_1^1|s_1^1)}{\widehat{p}(a_1^1)}- \dots \\
& \dots +\frac{1}{\beta} \log \frac{\hat{\pi}_\Fen(a_0^2|s')}{\widehat{p}(a_0^2)} - r(s_1^2, s', a_0^2) +\frac{1}{\beta} \log \frac{\hat{\pi}_\Fen(a_1^2|s_1^2)}{\widehat{p}(a_1^2)} - \dots \Big) = \\
& = \sum_{(a_0^1, s_1^1, a_1^1, \dots, a_0^2, s_1^2, a_1^2, \dots )} p(a_0^1, s_1^1, a_1^1, \dots, s', a_0^2, s_1^2, a_1^2, \dots, g|s) \cdot  \\ 
& \cdot \Big( \frac{1}{\beta} \log \frac{\pifi{1}(a_0^1|s)}{\widehat{p}(a_0^1)} - r(s_1^1,s,a_0^1) +\frac{1}{\beta} \log \frac{\pifi{1}(a_1^1|s_1^1)}{\widehat{p}(a_1^1)}- \dots \\
& \dots +\frac{1}{\beta} \log \frac{\pifi{2}(a_0^2|s')}{\widehat{p}(a_0^2)} - r(s_1^2, s', a_0^2) +\frac{1}{\beta} \log \frac{\pifi{2}(a_1^2|s_1^2)}{\widehat{p}(a_1^2)} - \dots \Big), \\
\end{aligned}
\label{eq:newF}
\end{equation} 

where in the last equation we substituted $\hat{\pi}_\Fen$ by $\pifi{1}$
or $\pifi{2}$ as per the definition of $\hat{\pi}_\Fen$. Now, we can use the linearity of the expected value, split the sum in (\ref{eq:newF}) and marginalizing away the superfluous variables to obtain the individual free energies $\Fen^{\pifi{1}}_{s'}(s)$ and $\Fen^{\pifi{2}}_g(s')$,

\small
\begin{equation}
\begin{aligned}
& \Fen^{\hat{\pi}_\Fen}_g(s) = \\
& = \underbrace{\sum_{(a_0^1, s_1^1, a_1^1, \dots)} p(a_0^1, s_1^1, a_1^1, \dots, s'|s) \left(\frac{1}{\beta} \log \frac{\pifi{1}(a_0^1|s)}{\widehat{p}(a_0^1)} - r(s_1^1, s, a_0^1) +\frac{1}{\beta} \log \frac{\pifi{1}(a_1^1|s_1^1)}{\widehat{p}(a_1^1)} - \dots \right)}_{\Fen^{\pifi{1}}_{s'}(s)} + \\
& \underbrace{\sum_{(a_0^2, s_1^2, a_1^2, \dots)} p(a_0^2, s_1^2, a_1^2, \dots, g|s') \left(\frac{1}{\beta} \log \frac{\pifi{2}(a_0^2|s')}{\widehat{p}(a_0^2)} - r(s_1^2, s', a_0^2) +\frac{1}{\beta} \log \frac{\pifi{2}(a_1^2|s_1^2)}{\widehat{p}(a_1^2)} - \dots \right)}_{\Fen^{\pifi{2}}_g(s')} = \\
& \Fen^{\pifi{1}}_{s'}(s) + \Fen^{\pifi{2}}_g(s')  < \Fen^{\pif}_g(s),\\
\end{aligned}
\label{eq:inequality_expansion}
\end{equation} 
\normalsize

The last inequality directly comes from our main assumption of
violating the triangular inequality.  In equation
(\ref{eq:inequality_expansion}) we have proven that
$\Fen^{\hat{\pi}_\Fen}_g(s) < \Fen^{\pif}_g(s)$; i.e.\ when moving from
$s$ to $g$ via $s'$ and the composite policy $\hat{\pi}_\Fen$, the free energy is smaller than the free
energy obtained using  $\pif$. But $\pif$ was assumed optimal, hence
the assumption that the triangle inequality is violated leads to a
contradiction. It therefore must hold and the proof is established.
$\square$

\nolinenumbers

\bibliography{library}

\bibliographystyle{abbrvnat} 

\end{document}